\documentclass[runningheads]{llncs}
\usepackage[T1]{fontenc}
\usepackage{booktabs}
\usepackage{amsmath}
\usepackage{amssymb}
\usepackage{multirow}
\usepackage{float}
\usepackage{fancyvrb}
\usepackage{listings}
\usepackage{tcolorbox}
\usepackage{verbatim}
\usepackage{mdframed}
\usepackage[pagebackref,breaklinks,colorlinks]{hyperref}
\usepackage{graphicx,verbatim}
\definecolor{lightred}{RGB}{250, 203, 204}
\definecolor{lightyellow}{RGB}{254, 250, 217}
\definecolor{lightgreen}{RGB}{204, 255, 203}

\usepackage{colortbl}
\definecolor{midgreen}{rgb}{0.0, 0.75, 0.0}

\begin{document}
\title{M$^3$Builder: A Multi-Agent System for Automated Machine Learning in Medical Imaging}

%

\author{\setlength{\baselineskip}{14pt}Jinghao Feng\inst{*,1,2}  Qiaoyu Zheng\inst{*,1,2}  Chaoyi Wu\inst{1,2}  Ziheng Zhao\inst{1,2}  \\[2pt] Ya Zhang\inst{1,2}  Yanfeng Wang\inst{\dag,1,2} and Weidi Xie\inst{\dag,1,2}}

\authorrunning{Feng et al.}
\titlerunning{M$^3$BUILDER}
\institute{\setlength{\baselineskip}{14pt} $^{1}$Shanghai Jiao Tong University \hspace{1cm} $^{2}$Shanghai AI Laboratory \\ $^{*}$ Equal Contribution \hspace{2.5cm} $^{\dag}$ Corresponding Author}

\maketitle              
\begin{abstract}
Agentic AI systems have gained significant attention for their ability to autonomously perform complex tasks. However, 
their reliance on well-prepared tools limits their applicability in the medical domain, which requires to train specialized models. 
In this paper, we make three contributions:
\textbf{(i)} We present \textbf{M$^3$Builder}, a novel multi-agent system designed to automate machine learning (ML) in medical imaging. 
At its core, M$^3$Builder employs four specialized agents that collaborate to tackle complex, multi-step medical ML workflows, from automated data processing and environment configuration to self-contained auto debugging and model training.
These agents operate within a medical imaging ML workspace, 
a structured environment designed to provide agents with free-text descriptions of datasets, training codes, and interaction tools, enabling seamless communication and task execution.
\textbf{(ii)} To evaluate progress in automated medical imaging ML, 
we propose \textbf{M$^3$Bench}, a benchmark comprising four general tasks on 14 training datasets, across five anatomies and three imaging modalities, covering both 2D and 3D data.
\textbf{(iii)} We experiment with seven state-of-the-art large language models serving as agent cores for our system, such as Claude series, GPT-4o, and DeepSeek-V3. Compared to existing ML agentic designs, M$^3$Builder shows superior performance on completing ML tasks in medical imaging, \textbf{achieving a 94.29\% success rate using Claude-3.7-Sonnet as the agent core}, showing huge potential towards fully automated machine learning in medical imaging.


\end{abstract}
\section{Introduction}
Large Language Model (LLM)-powered agentic systems have demonstrated remarkable success across diverse domains, leveraging their ability to orchestrate specialized tools and solve complex, multi-step tasks with precision. 
However, their application in medical domain remains challenging, 
due to a shortage of well-prepared tools. 
Such challenge arises from two aspects: 
first, the complexity of medical workflows—spanning diverse diseases, imaging modalities, and task requirements—makes tool development and integration difficult; second, while clinicians are experts in their field, 
they often lack the coding skills or resources to create tools, widening the gap between clinical needs and tailored AI solutions.

\begin{figure}[t!]
\includegraphics[width=\textwidth]{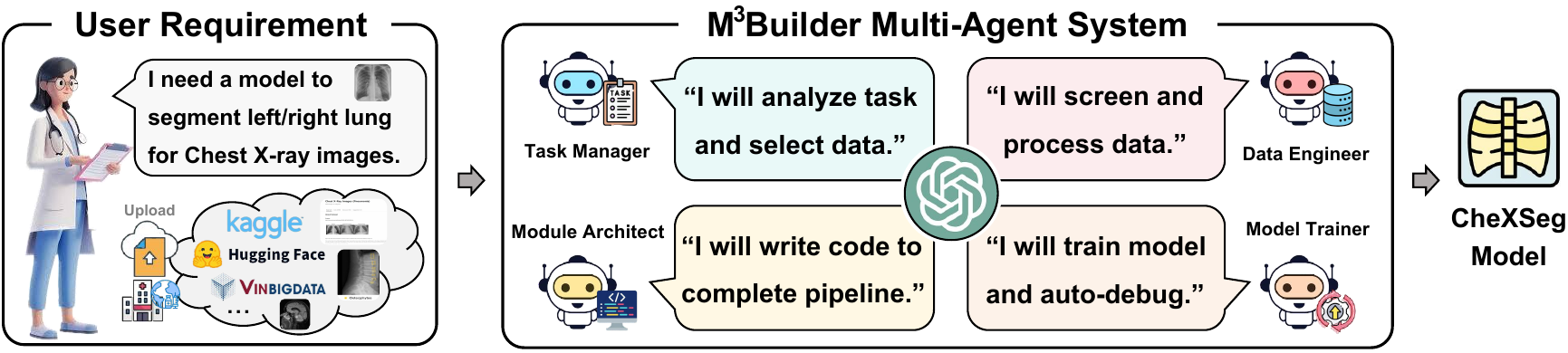}
\caption{General Pipeline of M$^3$Builder. Based on user requirements and available raw clinical datasets, four agents play different roles and collaborate to complete the procedure from task analysis, data processing to training model.}
\label{fig1}
\end{figure}

In this paper, we propose \textbf{M$^3$Builder} (M$^3$ stands for \textbf{M}utli-agent, \textbf{M}achine Learning, and \textbf{M}edical Imaging), 
a novel agentic framework for automating machine learning (ML) workflows in medical imaging. At its core, M$^3$Builder employs a multi-agent collaboration framework, 
where four specialized role-playing LLM agents work together to execute complex, multi-step workflows, collaboratively managing task requirement understanding, procedural planning, raw data processing, environment configuration, automated self-correction and model training execution.

Supporting this framework is the Medical Imaging ML workspace, 
a structured environment that provides the agents with initial datasets, pipeline templates, and interaction tools,
enabling seamless human-in-the-loop.
Together, the agents and the workspace form a cohesive system capable of autonomously developing AI tools for medical imaging analysis, 
bridging gaps in existing toolsets and facilitating the creation of a dynamic agentic ecosystem capable of evolving with minimal human intervention. 
Given a medical imaging-specific ML task and raw training data, \textbf{M$^3$Builder} autonomously manages the entire development process, 
from data preparation to model construction, training, and deployment.

In addition, we introduce a new benchmark, \textbf{M$^3$Bench}, designed to evaluate the performance for automated machine learning in medical imaging. \textbf{M$^3$Bench} comprises four general tasks: 
{\em organ segmentation}, {\em anomaly detection}, {\em disease diagnosis}, and {\em report generation}. These tasks are further associated with 14 detailed training datasets, spanning five anatomical regions and three primary imaging modalities, encompassing both 2D and 3D models. By covering a broad range of tasks and datasets, M$^3$Bench provides a standardized foundation for assessing the capabilities of agentic systems in automated ML in medical imaging.

Experimentally, we compare seven state-of-the-art (SOTA) LLMs as the agent core for M$^3$Builder, including GPT-4o~\cite{hurst2024gpt}, Claude-Sonnet series~\cite{TheC3}, and DeepSeek-V3~\cite{liu2024deepseek}.
Later, to validate the effectiveness of the proposed workspace and multi-agent collaboration framwork, we compare against other state-of-the-art ML agentic systems, including ML-AgentBench~\cite{huang2023mlagentbench}, Aider~\cite{aider}, Cursor Composer~\cite{cursor}, Windsurf Cascade~\cite{Windsurf} and Copilot Edits~\cite{copilot}. Our results demonstrate that \textbf{M$^3$Builder} consistently achieves superior performance across a range of metrics, advancing the state-of-the-art in fully autonomous ML workflows for medical imaging, \textbf{achieving a 94.29\% success rate using Claude-3.7-Sonnet as the agent core}, with 86.67\% on organ segmentation, 100\% on anomaly detection, 95\% on disease diagnosis, and 93.33\% on report generation.


\section{Method}
In this section, we present the proposed \textbf{M$^3$Builder} in detail, 
starting with the problem formulation, then a detailed description on the ML workspace initialization and the multi-agent collaboration framework.

\subsection{Problem Formulation}

\begin{figure}[t]
\includegraphics[width=\textwidth]{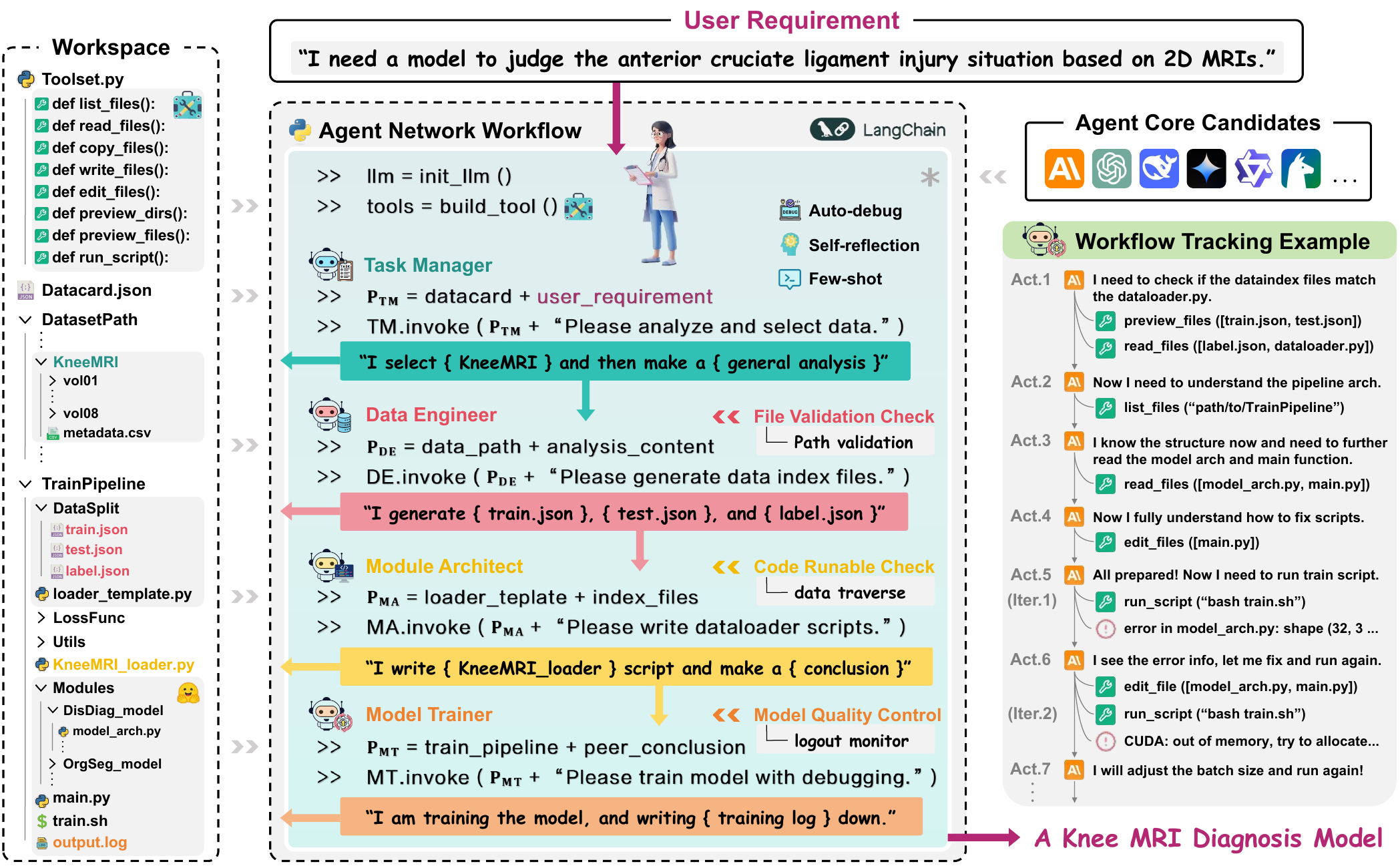}
\caption{Visualization of M$^3$Builder workflow: From user's free-text request to model delivery. The system integrates user requirements, a workspace with candidate data, tools, and code templates, a network of four specialized collaborative agents. A sample log tracks the Model Trainer agent's activities during diagnosis model development.} \label{fig2}
\vspace{-6pt}
\end{figure}

Given a task description on medical imaging analysis, denoted as $\mathcal{T}$, 
our objective is to automatically construct a functional AI model via multi-agents collaboration.  As shown in Fig.~\ref{fig2}, 
our proposed framework \textbf{M$^3$Builder} comprises two key components: 
the medical imaging ML workspace~($\mathcal{W}$), 
and the multi-agent collaboration framework~($\mathcal{A}$). 
Specifically, the workspace includes three elements: 
{\em data cards}, {\em toolset descriptions}, and {\em code templates}. 
The data cards are represented in natural language, while the toolset descriptions and code templates are provided in \texttt{Python}. Together, these elements form a structured `environment' that guides the automatic AI workflow.

Building on the workspace~($\mathcal{W}$), the multi-agent framework composes of four LLM agents with distinct roles, {\em i.e.}, $\mathcal{A} = \{a_1, a_2, a_3, a_4\}$.
These agents adopt a divide-and-conquer strategy to collaboratively address the complexities of the AI task. The framework iteratively performs code generation, editing and executing using tools defined by {\em toolset descriptions}, 
until a functional AI model is successfully produced. This process can be expressed as:
\begin{equation}
\{\mathcal{C}_{i}, \mathcal{R}_{i}\} = \mathcal{A}(\mathcal{C}_{i-1}, \mathcal{R}_{i-1}, \mathcal{T} \mid \mathcal{W}),
\end{equation}
where $\mathcal{C}_{i}$ denotes code and scripts generated or edited in the $i_{th}$ iteration, $\mathcal{R}_{i}$ denotes the compiler feedback in \texttt{Python} environment, with $\mathcal{C}_0 = \mathcal{R}_0 = \varnothing$.

In the following sections, we present more details for the AI workspace initialization and the multi-agent collaboration framework.

\subsection{Workspace Initialization}

The ML Workspace for medical imaging analysis is designed to serve three key purposes:
(i) it provides the multi-agent framework with metadata of the available datasets, enabling informed dataset selection;
(ii) it supplies initial code templates, offering a standardized starting point for agents and demonstrating a typical AI model training pipeline;
(iii) it defines and describes all tools available to the agents, restricting their action space to a predefined, complete set of operations.
In summary, the workspace equips the multi-agent framework with necessary resources for dataset preparation, a coding foundation, and a well-defined action space, corresponding to dataset descriptions, code templates, and toolset definitions.

\vspace{5pt} \noindent \textbf{Dataset Preparation.} 
The workspace includes a range of medical imaging datasets, each accompanied by a structured datacard for standardized descriptions. Each datacard contains key information such as the dataset name, a concise summary (covering data sources, categorical classifications, and scope), and detailed metadata. 
This flexible format ensures that any description meeting these criteria qualifies as a valid datacard, including documentation provided with the dataset itself. To enable seamless integration, users can add new datasets by completing corresponding datacards. Initially, the workspace provides 14 medical imaging datasets with their datacards as examples.

\vspace{5pt} \noindent \textbf{Code Template Design.}
To streamline AI model training while maintaining flexibility, 
we prepare a set of standardized code templates based on the Transformers Trainer framework and nnU-Net\cite{isensee2021nnu}. These templates are tailored to four primary medical imaging tasks: 
{\em disease diagnosis}, {\em organ segmentation}, {\em anomaly detection}, and {\em report generation}. Each task is implemented as a modular package with configurable components, such as the main forward architecture and network backbone options (2D/3D models). Shared features across tasks include a unified selection of loss functions, data augmentation strategies, training utilities, and architectural frameworks. By offering these templates, we reduce the complexity of free-form coding while preserving adaptability for diverse AI tasks.

\vspace{5pt} \noindent \textbf{Interaction Toolset.}
To facilitate the agentic system to interact with the PC environment, we develop a comprehensive interaction toolset, comprising eight functions, as shown in the top-right corner of Figure~\ref{fig2}. Inspired by MLAgentBench, the toolset includes: \textit{list\_files}, \textit{read\_files}, \textit{copy\_files}, \textit{write\_files}, \textit{edit\_files}, and \textit{run\_script} functions. 
In addition, we also introduce two specialized tools: 
\textit{preview\_dirs} and \textit{preview\_files}. 
The former processes large dataset directories containing numerous files that exceed the LLM's context window, while the latter extracts key segments from oversized metadata files that cannot be fully processed by the \textit{read\_files} function.


\subsection{Multi-Agent Collaboration Framework}

This section introduces our multi-agent collaboration framework~($\mathcal{A}$), which decomposes the AI task into four sub-tasks and assigns them to four specialized role-playing LLMs agents: Task Manager, Data Engineer, Module Architect, and Model Trainer, 
denoted as $\{a_1, a_2, a_3, a_4\}$, respectively, as illustrated in Fig.~\ref{fig2}. Each agent is responsible for a specific role, and together they collaborate iteratively to construct the final AI model. 
We utilize a set of system prompts to define the role and working logic of each agent, shown in supplementary files.

\vspace{5pt}\noindent \textbf{Task Manager}
acts as the coordinator of the framework. 
Its primary responsibilities include selecting the most suitable dataset for the task, or alternatively, asking users to upload raw datasets with associated datacard to the workspace as the supplementary of pre-existing datasets, and generating a comprehensive planning document $\mathcal{P}$ to guide the collaboration among the other agents. 
Specifically, given a user-provided task description, 
as exemplified by ``user requirements'' in Fig.~\ref{fig2}, the Task Manager will identify and select the optimal dataset~($\mathcal{D}$) for model training and generate the planning documents. 
This process can be formally represented as:
\begin{equation}
    \{\mathcal{D}, \hspace{2pt} \mathcal{P}\} = a_1(\mathcal{T} \mid \mathcal{W}_d),
\end{equation}
where $\mathcal{W}_d$ denotes the data card in the pre-defined workspace. 

\vspace{5pt}
\noindent \textbf{Data Engineer} 
is responsible for dataset preparation and processing. 
It transforms raw data into a format suitable for model training by performing tasks such as pre-screening the organizational structure of large-scale datasets, analyzing metadata files to extract relevant information, and splitting datasets into training and testing subsets. A critical aspect of the Data Engineer's role is its iterative interaction with the external compiler environment. It generates, edits, and refines code, incorporating feedback from the compiler, until the code executes successfully. This iterative process ensures the dataset preparation code is both robust and functional. The process can be expressed as:
\begin{equation}
    \{\mathcal{C}_{i},\mathcal{R}_{i}\} = a_2(\mathcal{C}_{i-1},\mathcal{R}_{i-1},\mathcal{T} \mid \mathcal{D}, \mathcal{P}),
\end{equation}
where $\mathcal{C}_i$ represents the $i^{th}$ version of the code generated by the Data Engineer, and $\mathcal{R}_{i}$ denotes the compiler feedback from the \texttt{Python} environment. Similarly as defined in \textbf{Problem Formulation}, $\mathcal{C}_0 = \mathcal{R}_0 = \varnothing$.

\vspace{5pt}
\noindent \textbf{Module Architect.}
Upon the dataset preparation, the Module Architect integrates essential components into the training pipeline, including developing dataloader scripts, designing appropriate model architecture, initializing the entrance function for training, 
and selecting other components, such as loss functions, data augmentation strategies, and training utilities. 
Notebly, the Module Architect will iteratively validates the dataloader to ensure it outputs batches with correct shapes and formats. The process can be formulated as:
\begin{equation}
    \{\mathcal{C}_{i},\mathcal{R}_{i}\} = a_3(\mathcal{C}_{i-1},\mathcal{R}_{i-1},\mathcal{T} \mid \mathcal{D}, \mathcal{W}_c),
\end{equation}
where $\mathcal{W}_c$ represents the code templates in the workspace. Similar to the Data Engineer, the code generation process in the Module Architect is also iterative and we denote the final iteration output as $\mathcal{C}_\text{MA}$. After the integration of these modules, the architect finally synthesizes a summary $\mathcal{S}$ of all completed work.

\vspace{5pt}
\noindent \textbf{Model Trainer} 
finalizes the debugging and optimizing the training procedure. 
Building upon the pipeline established by the Module Architect, the Model Trainer first verifies the completeness and correctness of the training framework thus far. 
It then selects hyperparameters to meet the model's specific training requirements, while retaining the authority to modify any part of the code—including model code, dataloader code, and training scripts—based on errors encountered during training. The entire workflow can be expressed as:
\begin{equation}
    \{\mathcal{C}_{i},\mathcal{R}_{i}\} = a_4(\mathcal{C}_{i-1},\mathcal{R}_{i-1},\mathcal{T} \mid \mathcal{D},\mathcal{C}_\text{MA}, \mathcal{S}).
\end{equation}
After iteratively performing the above code generation pipeline until successfully executed, the final code will produce a desired AI model. 

\section{Benchmark \& Experiments}
We evaluate \textbf{M$^3$Builder}, using 14 diverse ML model development tasks across 4 radiology domains, paired with matched clinical datasets to enable comprehensive testing. The evaluation employs 7 leading LLMs as agent cores: GPT-4~\cite{hurst2024gpt}, Claude-3.7-Sonnet\cite{TheC3}, Claude-3.5-Sonnet~\cite{TheC3}, DeepSeek-v3~\cite{liu2024deepseek}, Gemini-2.0-flash~\cite{team2023gemini}, Qwen-2.5-max~\cite{yang2024qwen2}, and Llama-3.3-70B~\cite{dubey2024llama}. 
Quantitatively, we assess the system's effectiveness through analysis on task completion , framework superiority, and agent role-specification.

\begin{table}[t!]
\caption{Task Completion Performance Across LLMs. Each experiment undergoes multi-runs, with results shown as successful completions over total rounds (a/b format). \colorbox{lightgreen}{Green} cells indicate that all runs passed, \colorbox{lightyellow}{Yellow} indicates partially passed, and \colorbox{lightred}{Red} indicates that all runs failed.}
\label{tab1}
\centering
\begin{tabular}{ll|ccccccc}
\toprule
                             &            & \multicolumn{7}{c}{\textbf{LLMs}}                                                         \\
\multirow{-2}{*}{\textbf{Task}} & \multirow{-2}{*}{\textbf{Dataset}} & Son.3.7 & Son.3.5 & GPT4o& QwMax& DeepSeekV3 & Gemini2& Llama3.3 \\ \midrule
                             & BTCV  &     \cellcolor[HTML]{ccffcb}5/5 & \cellcolor[HTML]{fefad9}4/5& \cellcolor[HTML]{ccffcb}5/5 & \cellcolor[HTML]{fefad9}3/5 & \cellcolor[HTML]{fefad9}2/5 & \cellcolor[HTML]{facbcc}0/5& \cellcolor[HTML]{fefad9}1/5\\
                             & Verse  &    \cellcolor[HTML]{fefad9}4/5& \cellcolor[HTML]{fefad9}4/5& \cellcolor[HTML]{ccffcb}5/5 &  \cellcolor[HTML]{fefad9}2/5     &       \cellcolor[HTML]{facbcc}0/5 &       \cellcolor[HTML]{facbcc}0/5&       \cellcolor[HTML]{facbcc}0/5\\
\multirow{-3}{*}{OrgSeg}     & OASIS  &    \cellcolor[HTML]{fefad9}4/5& \cellcolor[HTML]{fefad9}3/5& \cellcolor[HTML]{fefad9}3/5& \cellcolor[HTML]{fefad9}1/5      &       \cellcolor[HTML]{fefad9}1/5&       \cellcolor[HTML]{facbcc}0/5&       \cellcolor[HTML]{facbcc}0/5\\ \midrule
                             & COV19  &    \cellcolor[HTML]{ccffcb}5/5 & \cellcolor[HTML]{ccffcb}5/5 & \cellcolor[HTML]{fefad9}4/5 &   \cellcolor[HTML]{fefad9}3/5    &       \cellcolor[HTML]{fefad9}1/5 &       \cellcolor[HTML]{fefad9}1/5&       \cellcolor[HTML]{facbcc}0/5\\
                             & INS22 &     \cellcolor[HTML]{ccffcb}5/5 & \cellcolor[HTML]{ccffcb}5/5 & \cellcolor[HTML]{fefad9}4/5 &   \cellcolor[HTML]{fefad9}2/5    &       \cellcolor[HTML]{fefad9}2/5 &       \cellcolor[HTML]{facbcc}0/5&       \cellcolor[HTML]{facbcc}0/5\\
                             & Panc &      \cellcolor[HTML]{ccffcb}5/5 & \cellcolor[HTML]{ccffcb}5/5 & \cellcolor[HTML]{ccffcb}5/5 &   \cellcolor[HTML]{fefad9}4/5    &       \cellcolor[HTML]{fefad9}2/5 &       \cellcolor[HTML]{facbcc}0/5&       \cellcolor[HTML]{fefad9}1/5\\
\multirow{-4}{*}{AnoDet}     & XDet10 & \cellcolor[HTML]{ccffcb}5/5 & \cellcolor[HTML]{ccffcb}5/5 & \cellcolor[HTML]{ccffcb}5/5 &   \cellcolor[HTML]{fefad9}3/5    &       \cellcolor[HTML]{fefad9}1/5 &       \cellcolor[HTML]{fefad9}1/5&       \cellcolor[HTML]{facbcc}0/5\\ \midrule
                             & ADNI &      \cellcolor[HTML]{ccffcb}5/5 & \cellcolor[HTML]{fefad9}4/5 & \cellcolor[HTML]{fefad9}2/5 &  \cellcolor[HTML]{fefad9}1/5     &       \cellcolor[HTML]{fefad9}1/5&       \cellcolor[HTML]{facbcc}0/5&       \cellcolor[HTML]{facbcc}0/5\\
                             & KneeMR &    \cellcolor[HTML]{fefad9}4/5 & \cellcolor[HTML]{ccffcb}5/5 & \cellcolor[HTML]{fefad9}4/5 &  \cellcolor[HTML]{fefad9}2/5   &       \cellcolor[HTML]{fefad9}1/5&       \cellcolor[HTML]{facbcc}0/5&       \cellcolor[HTML]{facbcc}0/5\\
                             & CC-CCII &   \cellcolor[HTML]{ccffcb}5/5 & \cellcolor[HTML]{ccffcb}5/5 & \cellcolor[HTML]{ccffcb}5/5 &  \cellcolor[HTML]{facbcc}0/5    &       \cellcolor[HTML]{facbcc}0/5&       \cellcolor[HTML]{facbcc}0/5&       \cellcolor[HTML]{facbcc}0/5\\
\multirow{-4}{*}{DisDiag}    & KidCT  &    \cellcolor[HTML]{ccffcb}5/5 & \cellcolor[HTML]{ccffcb}5/5 & \cellcolor[HTML]{fefad9}4/5 &  \cellcolor[HTML]{fefad9}2/5     &       \cellcolor[HTML]{fefad9}1/5 &       \cellcolor[HTML]{facbcc}0/5&       \cellcolor[HTML]{facbcc}0/5\\ \midrule
                             & CT-RATE &   \cellcolor[HTML]{ccffcb}5/5& \cellcolor[HTML]{fefad9}4/5 & \cellcolor[HTML]{fefad9}4/5 &  \cellcolor[HTML]{fefad9}2/5     &       \cellcolor[HTML]{fefad9}1/5 &       \cellcolor[HTML]{facbcc}0/5&       \cellcolor[HTML]{facbcc}0/5\\
                             & GenomBra & \cellcolor[HTML]{ccffcb}5/5& \cellcolor[HTML]{ccffcb}5/5 & \cellcolor[HTML]{fefad9}4/5 &  \cellcolor[HTML]{fefad9}3/5     &       \cellcolor[HTML]{fefad9}1/5 &       \cellcolor[HTML]{facbcc}0/5&       \cellcolor[HTML]{fefad9}1/5\\
\multirow{-3}{*}{RepGene}    & IU-Xray &   \cellcolor[HTML]{fefad9}4/5 & \cellcolor[HTML]{fefad9}4/5 & \cellcolor[HTML]{fefad9}3/5 &  \cellcolor[HTML]{fefad9}1/5     &       \cellcolor[HTML]{fefad9}2/5&       \cellcolor[HTML]{facbcc}0/5&       \cellcolor[HTML]{facbcc}0/5\\ \midrule
\multicolumn{2}{c|}{\textbf{Average(\%)}} & \textbf{94.29}& 90.00& 81.43                       &   41.43    &   22.86    &       4.29   &   4.29   \\ \bottomrule
\end{tabular}
\vspace{-6pt}
\end{table}

\subsection{Task Definition \& Data Preparation}
In this paper, we experiment with 14 tasks spanning \textit{organ segmentation}, \textit{anomaly detection}, \textit{disease diagnosis}, and \textit{report generation}. These tasks are systematically categorized by anatomic regions (head \& neck, chest, abdomen \& pelvis, limb, spine), imaging modalities (X-ray, CT, MRI), and dimensionality (2D/3D). Each task is precisely defined, for instance: ``please build a model for covid-19/pneumonia classification from 3D chest CT images.'' 

To facilitate model development, we have curated 14 domain-specific datasets: ADNI~\cite{Jack2008TheADNI}, KneeMRI~\cite{Stajduhar2017-au}, CC-CCII~\cite{CCCCII}, CT-Kidney~\cite{ctkidney}, BTCV~\cite{BTCV}, MSD Pancreas~\cite{MSD}, VerSe~\cite{verse1,verse2,verse3}, L2R-OASIS~\cite{l2roasis}, COVID-19~\cite{covid19}, CT-RATE~\cite{ctrate}, INSTANCE2022~\cite{instance22}, ChestX-Det10~\cite{chestxdet10}, RadGenome-Brain-MRI~\cite{lei2024autorg}, and IU-Xray~\cite{iuxray}. Each dataset is accompanied by a datacard with metadata from their original publications. Additionally, we expand the datacard pool through GPT-generated synthetic entries to simulate real-world scenarios where suitable datasets must be selected from large-scale repositories for model training.

\subsection{Task Completion Analysis}

We design a specific model-building task for each dataset based on its unique characteristics, resulting in a total of 14 tasks. 
For each task, seven large language models (LLMs) are run independently five times, with a maximum of 100 actions~(tool invocations) allowed per execution. The task completion is defined as the successful training of a model with performance on the test set falling within an acceptable range. As the results shown in Tab.\ref{tab1}, the performance of different LLMs exhibit significant variation, with Claude-3.7-Sonnet achieving the highest completion rate of 94.29\%, while Gemini2.0 and Llama3.3 only reach 4.29\%. The experimental results demonstrate 
our M$^3$Builder possesses impressive and robust capabilities for medical imaging model training automation, particularly when leveraging LLMs that excel in tool utilization and code generation, which significantly enhances the system's overall performance.

\subsection{Compare to State-of-the-art Agentic Systems}
We compare M$^3$Builder with other agentic systems including MLAgent-Bench, Aider, Cursor Composer, Windsurf Cascade, and Copilot Edits (all using Sonnet as the agent core). Each system performed each task twice on our workspace under their built-in framework. As shown in Tab.~\ref{tab2}, across radiology tasks (Organ Segmentation, Anomaly Detection, Disease Diagnosis, and Report Generation), MLAgent-Bench performed poorly due to insufficient data structure understanding capabilities. Other frameworks achieved only moderate success rates (39.29\% max) due to single-agent limitations, required human-in-the-loop confirmation, increasing operational complexity and max iteration constraints. Our M$^3$Builder demonstrated superior performance with a 42.85\% higher average success rate while requiring fewer action steps and execution iterations.

\begin{table}[t!]
\caption{Framework Comparison with SOTAs and Ablations on System Design using Sonnet. Results are averaged over two runs per task in dataset-level. ``w/o Colab'' represents single-agent execution, and ``Iters'' means the self-correction rounds.}
\label{tab2}
\centering
\begin{tabular}{l|ccccc|cccc}
\toprule
              & \multicolumn{5}{c|}{\textbf{Completion Runs (Total) ~$\uparrow$}}                             & \multicolumn{4}{c}{\textbf{Average Actions (Iters)~$\downarrow$}} \\
\multirow{-2}{*}{\textbf{\begin{tabular}[c]{@{}l@{}}Agentic\\ System\end{tabular}}} &
  Seg. &
  Det. &
  Diag. &
  Gen. &
  Avg(\%) &
  Seg. &
  Det. &
  Diag. &
  Gen. \\ \midrule
MLA-Bench     & 0(6) & 0(8) & 0(8)                         & 0(6) & 0.00                          & -(-)        & -(-)         & -(-)       & -(-)       \\
Aider &
  \cellcolor[HTML]{D8F1F5}2(6) &
  3(8) &
  2(8) &
  \cellcolor[HTML]{D8F1F5}3(6) &
  35.71 &
  58.0(6.5) &
  51.3(7.0) &
  44.5(4.5) &
  56.7(5.7) \\
Cursor Comp   & 1(6) & 3(8) & 3(8)                         & 2(6) & 32.14                         & 42.0(7.0)   & 35.7(4.3)    & 36.3(4.0)  & 51.5(5.5)  \\
Wsurf Casc    & 1(6) & 3(8) & \cellcolor[HTML]{D8F1F5}4(8) & 2(6) & 35.71                         & 39.0(5.0)   & 36.0(4.3)    & 35.3(4.8)  & 48.5(5.5)  \\
Copilot Edits &
  \cellcolor[HTML]{D8F1F5}2(6) &
  \cellcolor[HTML]{D8F1F5}4(8) &
  3(8) &
  2(6) &
  \cellcolor[HTML]{D8F1F5}39.29 &
  48.0(3.5) &
  45.3(3.5) &
  44.7(4.0) &
  46.5(4.5) \\ 
\textbf{Ours} & \cellcolor[HTML]{A8DBE4}4(6) & \cellcolor[HTML]{A8DBE4}7(8) & \cellcolor[HTML]{A8DBE4}8(8)                         & \cellcolor[HTML]{A8DBE4}4(6) & \cellcolor[HTML]{A8DBE4}82.14 & 34.5(1.8)   & 25.29(2.4)   & 35.0(1.9)  & 32.0(2.3)  \\ \midrule
w/o Colab     & 3(6) & 3(8) & 3(8)                         & 2(6) & 39.29                         & 33.6(4.7)   & 37.5(4.8)    & 33.3(4.3)  & 33.5(3.5)  \\
w/o Debug     & 2(6) & 3(8) & 1(8)                         & 0(6) & 21.43                         & 30.5(1.0)   & 24.0(1.0)    & 33.0(1.0)  & -(-)       \\
w/o Reflect   & 4(6) & 7(8) & 7(8)                         & 4(6) & 78.57                         & 28.3(2.3)   & 24.9(4.1)    & 35.4(3.9)  & 41.3(5.8)  \\
w/o Fewshot   & 3(6) & 4(8) & 6(8)                         & 3(6) & 57.14                         & 37.3(4.0)   & 23.5(4.3)    & 32.3(4.1)  & 34.0(3.0)  \\ \bottomrule
\end{tabular}
\vspace{-6pt}
\end{table}

\subsection{Ablation Study}

Here we present ablation studies on our system design, examining the impact of: single-agent versus multi-agent collaboration, auto-debugging capability, self-reflection mechanisms, and workflow few-shot examples. Results in Tab.~\ref{tab2} indicate that self-reflection has minimal influence on system performance, while auto-debugging proves crucial for successful training. Multi-agent collaboration and well-crafted example instructions also significantly impact performance, with their absence resulting in 42.85\% and 25.00\% performance gaps, respectively.


\subsection{Analysis on Different Agent Roles}

We evaluate the performance of each role-specific agent in \textbf{M$^3$Builder}, using distinct success criteria: 
The Task Managers is assessed on its ability to select appropriate training datasets, the Data Engineer on generating valid data index files with correct structures and paths, the Module Architect on producing executable scripts for data loading, and the Model Trainer on successfully completing model training.

\begin{table}[t!]
\caption{Role-specific agent performance on tasks using Sonnet. ``Run'', ``Act'', ``Iter'' and ``Tkn'' respectively denote ``execution rounds'', ``actions'', ``iterations'' and ``tokens''.}\label{tab3}
\vspace{-3pt}
\centering
\footnotesize
\begin{tabular}{l|cccc|cccc|cccc|cccc}
\toprule
\multirow{2}{*}{\textbf{Task}} &
  \multicolumn{4}{c|}{\textbf{Task Manager}} &
  \multicolumn{4}{c|}{\textbf{Data Engineer}} &
  \multicolumn{4}{c|}{\textbf{Module Architect}} &
  \multicolumn{4}{c}{\textbf{Model Trainer}} \\
        & Run & Act & Iter & Tkn  & Run & Act  & Iter & Tkn & Run & Act  & Iter & Tkn & Run & Act  & Iter & Tkn  \\ \midrule
Seg.  & 6/6 & 1.3 & 1.0  & 4.3k & 5/6 & 10.2 & 1.3  & 72k & 5/6 & 10.5 & 2.3  & 74k & 4/6 & 10.7 & 3.3  & 115k \\
Det.  & 8/8 & 2.0 & 1.0  & 4.8k & 8/8 & 10.0 & 1.4  & 92k & 7/8 & 10.6 & 2.1  & 61k & 7/8 & 9.3  & 2.1  & 67k  \\
Diag. & 8/8 & 2.0 & 1.0  & 4.4k & 8/8 & 8.9  & 1.4  & 116k & 7/8 & 11.3 & 2.0  & 84k & 8/8 & 9.3  & 2.0  & 198k \\
Gen.  & 6/6 & 2.0 & 1.0  & 4.2k & 6/6 & 8.7  & 1.3  & 66k  & 5/6 & 10.8 & 2.5  & 91k & 5/6 & 11.2 & 2.2  & 70k  \\ \bottomrule
\end{tabular}
\vspace{-9pt}
\end{table}

As shown in Tab. \ref{tab3}, the Task Manager demonstrated exceptional accuracy in task analysis, with stable token usage across all tasks.
In contrast, the other three agent roles show greater variability in token consumption and execution attempts. This variability stems from the strict requirements for code organization, data preprocessing, and achieving error-free training within five iterations. Despite these challenges, the majority of agents successfully completed or nearly completed all assigned task executions, showcasing the robustness and adaptability of the multi-agent framework.

\section{Conclusion}
In this paper, we present M$^3$Builder, an agentic system for automating machine learning in medical imaging tasks. Our approach combines an efficient medical imaging ML workspace with  free-text descriptions of datasets, code templates, and interaction tools. Additionally, we propsoe a multi-agent
collaborative agent system designed specifically for AI model building, with four role-playing LLMs, Task Manager, Data Engineer, Module Architect, and Model Trainer. In benchmarking against five SOTA agentic systems across 14 radiology task-specific datasets, M$^3$Builder achieves a 94.29\% model building success rate with Claude-3.7-Sonnet standing out among seven SOTA LLMs. Future work will extend beyond medical imaging to broader medical tasks, develop more robust tool-building agent systems, implement automated dataset preparation capabilities, and incorporate visual processing to better approximate clinical expertise.


\bibliographystyle{splncs04}
\bibliography{reference}

\clearpage
\appendix
\section{Supplementary Materials}

\subsection{Case Study.}
\begin{figure}
\includegraphics[width=\textwidth]{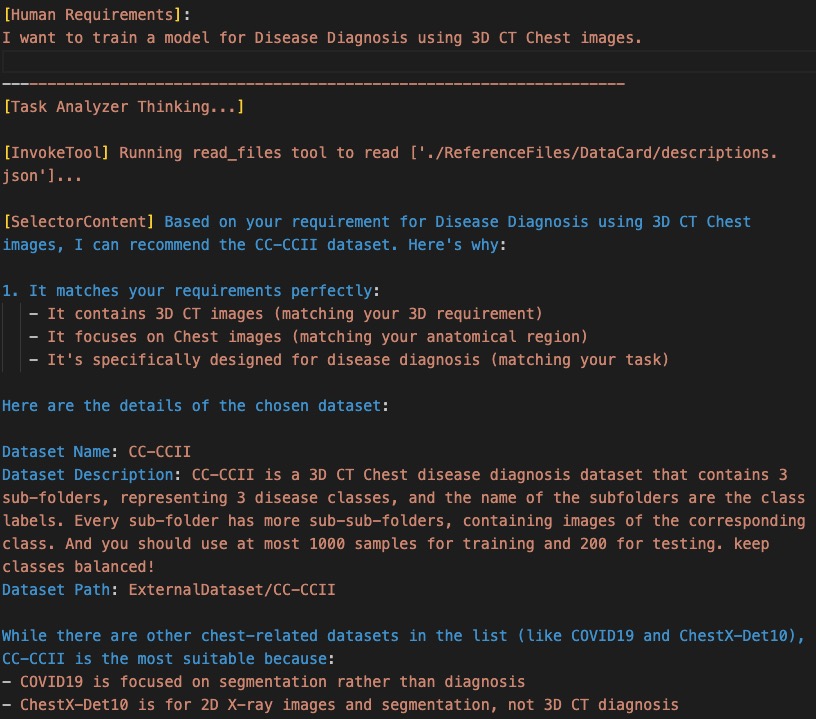}
\caption{Case Study: Terminal log of Task Analyzer.} \label{casefig1}
\end{figure}

Here we present terminal logs of a successful running case. In this case, we require the system to train a model for Disease Diagnosis using 3D CT Chest image data. As shown in Fig.~\ref{casefig1}, Task Analyzer first read thoroughly through all the dataset descriptions stored in $descriptions.py$, then choose the CC-CCII dataset and list the reasons of the choice. Task Analyzer also returns the detail information of CC-CCII dataset for other role agents.

\begin{figure}
\includegraphics[width=\textwidth]{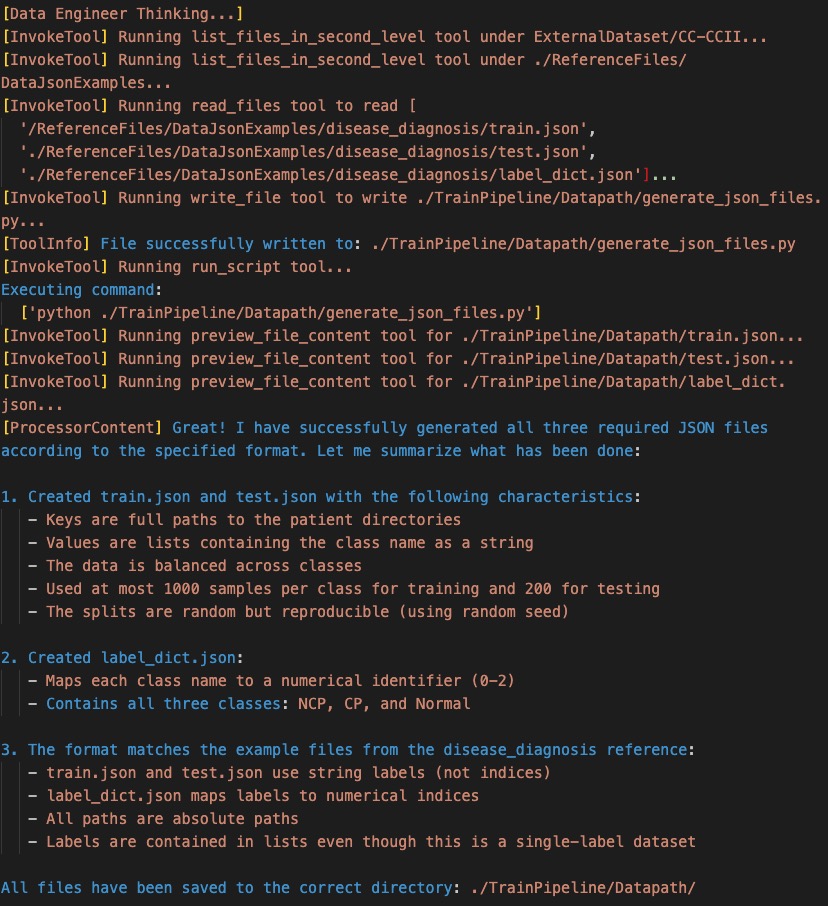}
\caption{Case Study: Terminal log of Data Engineer.} \label{casefig2}
\end{figure}

Following Task Analyzer, Data Engineer traverse the dataset files, splits the dataset into train/test sets. It refers to the example data index files, and generates a python script to generate the json files. The process is shown in Fig.~\ref{casefig2}. Finally, Data Engineer concludes its work, and shows a briefing to the user. 

\begin{figure}
\includegraphics[width=\textwidth]{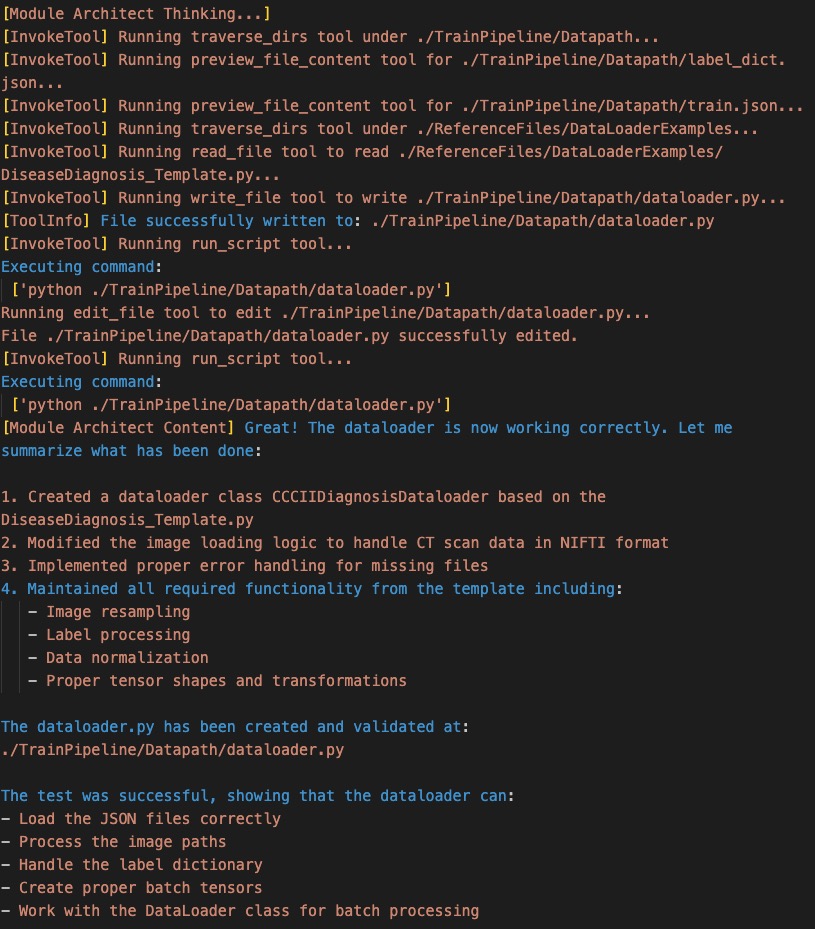}
\caption{Case Study: Terminal log of Module Architect.} \label{casefig3}
\vspace{-6pt}
\end{figure}

\vspace{5pt}
As shown in Fig.~\ref{casefig3}, Module Architect first checks that Data Engineer has generated all required files, then refers to the example dataloader scripts, and generates its own dataloader file, which guarantees standardized data processing and formatting. Before ending, it executes the generated dataloader file to check its correctness, by traversing and loading all the training and testing samples. Finally, it prints a briefing to conclude its work.

\vspace{5pt}

\begin{figure}
\includegraphics[width=\textwidth]{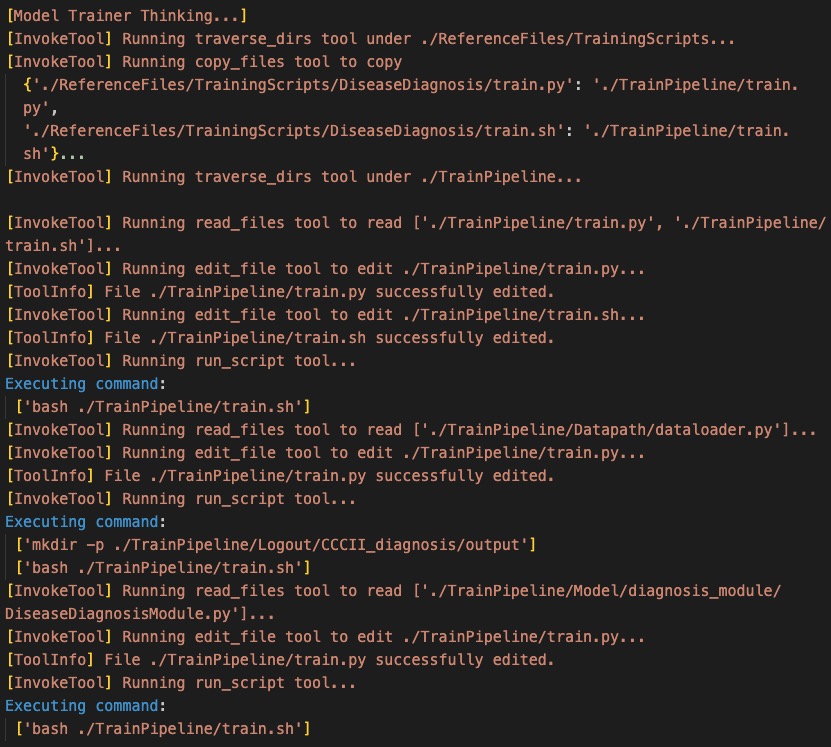}
\caption{Case Study: Terminal log of Model Trainer.} \label{casefig4}
\vspace{-6pt}
\end{figure}

Finally, Fig.~\ref{casefig4} shows the workflow of Model Trainer. It first choose a training code framework, and a ML module according to the medical task, then assemble the dataloader, the ML module and the training codes to form an executable training script. As we can see in 
Fig.~\ref{casefig4}, the first training attempts fails. After checking the trace-back information(not shown in the figure), it modifies the training scripts accordingly, and finally performs a successful run.

\subsection{Details about Inclusion Datasets}
We curated 14 datasets supporting different medical imaging tasks. Each dataset is unzipped in our workspace without further processing.

\begin{table}[]
\caption{Information of our initialized 14 datasets, including supported task, anatomy, modality and dimensions.}\label{tab3}
\vspace{-3pt}
\centering
\footnotesize
\begin{tabular}{l|llcc}
\toprule
\textbf{Dataset} & \textbf{Task} & \textbf{Anatomy} & \textbf{Modality} & \textbf{Dimension} \\ \midrule
BTCV         & Organ Segmentation & Abdom \& Pelvis & CT    & 3D \\
VerSe        & Organ Segmentation & Spine           & CT    & 3D \\
L2R-OASIS    & Organ Segmentation & Head \& Neck    & MRI   & 3D \\
COVID19      & Anomaly Detection  & Chest           & CT    & 3D \\
INSTANCE2022 & Anomaly Detection  & Head \& Neck    & CT    & 3D \\
MSD Pancreas & Anomaly Detection  & Abdom \& Pelvis & CT    & 3D \\
ChestX-Det10 & Anomaly Detection  & Chest           & X-ray & 2D \\
ADNI         & Disease Diagnosis  & Head \& Neck    & MRI   & 3D \\
KneeMRI      & Disease Diagnosis  & Limb            & MRI   & 2D \\
CC-CCII      & Disease Diagnosis  & Chest           & CT    & 3D \\
CT-Kidney    & Disease Diagnosis  & Abdom \& Pelvis & CT    & 2D \\
CT-RATE      & Report Generation  & Abdom \& Pelvis & CT    & 3D \\
BrainGenome  & Report Generation  & Head \& Neck    & X-ray & 3D \\
IU\_Xray     & Report Generation  & Chest           & X-ray & 2D \\ \bottomrule
\end{tabular}
\vspace{-9pt}
\end{table}

Each dataset is paired with a structured description in the datacard. Once a new dataset is uploaded into workspace, the corresponding data description should be inserted into this dynamic datacard file. The initialization of the datacard is shown as bellow:

\begin{mdframed}[backgroundcolor=lightgray!20]
\label{stage2_prompts}
\vspace{3pt}
\textbf{Datacard.json}
\vspace{5pt}
\hrule
\begin{verbatim}

[
    {
        "dataset name": "ADNI",
        "dataset description": "ADNI is a 3D MRI disease 
        diagnosis dataset that contains 1 subfolder and 1 csv 
        file. the subfolder(named ADNI_PT) contains images(in 
        .pt format), while the csv file contains lines of sample 
        info, where you should focus mainly on 2 columns: 
        'Image Data ID' and 'Group'(which is the class label). 
        Remember, only samples with these labels should be used: 
        ['AD','MCI','CN']! And you should use at most 1000 
        samples for training and 200 for testing. keep classes 
        balanced!",
        "dataset path": "/path/to/ADNI_Dataset"
    },
    {
        "dataset name": "KneeMRI",
        "dataset description": "KneeMRI is a 2D MRI disease 
        diagnosis dataset that contains 1 subfolder and 1 csv 
        file. the subfolder1 contains images(in .pck format), 
        while the csv file contains lines of sample info, where 
        you must focus mainly on 2 colomns: 'volumeFilename' 
        and 'aclDiagnosis'(which is the class label, could be 
        0 or 1 or 2). And you should use at most 1000 samples
        for training and 200 for testing. keep classes 
        balanced!",
        "dataset path": "/path/to/KneeMRI"
    },
    {
        "dataset name": "CC-CCII",
        "dataset description": "CC-CCII is a 3D CT disease 
        diagnosis dataset that contains 3 sub-folders, 
        representing 3 disease classes, and the name of the 
        subfolders are the class labels. Every sub-folder has 
        more sub-sub-folders, containing images of the 
        corresponding class. And you should use at most 1000 
        samples for training and 200 for testing. keep classes 
        balanced!",
        "dataset path": "/path/to/CC-CCII"
    },
    
    ...
    
    {
        "dataset name": "OASIS",
        "dataset description":"OASIS is a 3D Head & Neck MRI 
        organ segmentation dataset that contains 2 subfolders 
        - namely images and masks - containing nii.gz format 
        images and corresponding organ masks, and 1 json file: 
        labels.json, containing an ordered label list (please 
        do get a complete label_list from it).",
        "dataset path": "/path/to/OASIS"
    }
]

\end{verbatim}
\end{mdframed}

\subsection{Details about the toolset}
Here we provide each tool function in the toolset:
\begin{itemize}
    \item \textbf{list\_files}: Recursively scans the specified directory to identify and return paths of all code files (supporting common programming extensions like .py, .java, .cpp, etc.). Features intelligent directory skipping by automatically ignoring directories containing more than 1,000 files to prevent processing excessively large file collections. Returns a newline-separated string of file paths for further processing.
    \item \textbf{read\_files}: Opens and reads the entire content of a specified file, returning the complete text as a string. Supports UTF-8 encoding, making it suitable for examining source code, configuration files, or any text document. Essential for code analysis and file inspection tasks without modifying the original content.

    \item \textbf{copy\_files}: Creates an exact duplicate of a single file from a source location to a destination path. Automatically generates any necessary directory structure at the destination if it doesn't already exist. Preserves file metadata like timestamps and permissions using shutil.copy2, ensuring the duplicate maintains the characteristics of the original file.

    \item \textbf{write\_files}: Generates a new file with specified content at the designated file path. Automatically creates all necessary parent directories if they don't exist, ensuring the file can be written even to previously non-existent paths. Particularly useful for programmatically creating new script files, configuration files, or saving processed data.
    
    \item \textbf{edit\_files}: Completely overwrites an existing file with new content, replacing the original data entirely. Designed for direct file modification without the need to manually open and edit files. Critical for automated code refactoring, text transformation, or updating configuration files with revised settings in batch operations.
    
    \item \textbf{run\_script}: Executes shell commands in the operating system environment and captures their output. Leverages the ShellTool from LangChain to safely run commands and collect results. Enables interaction with the system shell to perform operations like running programs, executing system utilities, or triggering external processes from within the application.
    
    \item \textbf{preview\_dirs}: Performs a detailed analysis of a directory's structure by examining each immediate subfolder. For each entry, counts the total number of files and lists up to 100 file paths in natural sort order. Returns a structured dictionary with comprehensive information about directories and files, facilitating efficient navigation of complex file systems while limiting output size for large directories.
    
    \item \textbf{preview\_files}: Provides intelligent content summaries of structured and unstructured data files. For CSV files, displays the first 5 rows and total row count; for JSON, shows the first 5 key-value pairs or elements and total count; for text files, presents the first 10,000 words and total word count. Enables rapid content assessment without loading entire large files into memory, particularly valuable for data exploration tasks.
\end{itemize}

\subsection{Details about the Agent structure}
\begin{figure}[!]
\includegraphics[width=\textwidth]{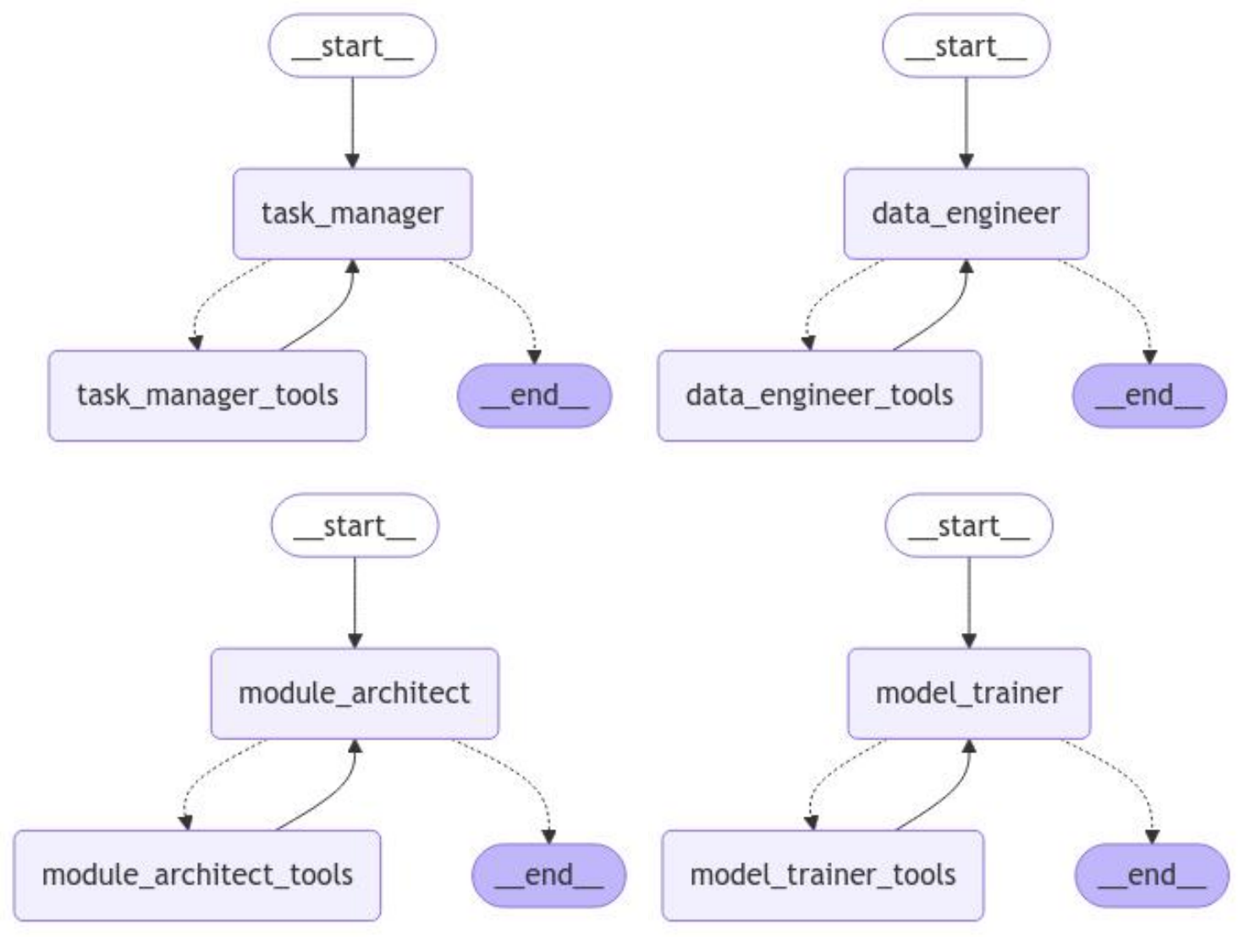}
\caption{The inner structure graph of our agents: Task Manager, Data Engineer, Module Architect and Model Trainer. All of them have the ability for tool using and will keep debugging until task completed.} \label{fig3}
\end{figure}

We use langgraph architecture for agent building and workflow graph compiling. As shown in Fig.~\ref{fig3}, All the agents have their own toolsets, which are subset of our proposed toolset containing 8 tools because of their specification and meaningless redundant information provided. The function calling loop and debugging mechanism ensure the task completion performance.

\subsection{Details about the Agent Core Candidates}
We select 7 SOTA LLMs for comparison. Most of them are closed-source model which are usually more powerful.

\begin{itemize}
    \item \textbf{Claude-3.7-Sonnet} -- Anthropic's latest model released in February 2025, featuring significant advancements in reasoning capabilities, contextual understanding, and tool utilization. This model demonstrates exceptional performance in complex multi-step reasoning tasks while maintaining high computational efficiency. Claude-3.7-Sonnet exhibits particularly strong capabilities in understanding nuanced instructions and maintaining coherence across lengthy interactions, making it ideal for our complex evaluation scenarios. The model api we use is: ``claude-3-7-sonnet-20250219''.
    
    \item \textbf{Claude-3.5-Sonnet} -- Released by Anthropic in 2024, this model represents a critical milestone in the Claude series, balancing performance and efficiency. \textit{We selected this model as the foundation for all our ablation studies due to its stable performance characteristics and consistent behavior across various experimental conditions. This strategic choice allowed us to isolate and measure the impact of individual components in our framework while maintaining a reliable baseline.} The model excels in reasoning tasks requiring detailed comprehension and precise execution of instructions. The model api we use is: ``claude-3-5-haiku-20241022''.
    
    \item \textbf{GPT-4o} -- OpenAI's advanced multimodal model that seamlessly integrates sophisticated vision capabilities with powerful language understanding and generation. This model demonstrates remarkable versatility across domains and task types, with particularly strong performance in scenarios requiring cross-modal reasoning. Its ability to process both textual and visual information makes it valuable for our evaluation of real-world applications where multimodal understanding is essential. The model api we use is: ``gpt-4o-2024-11-20''.
    
    \item \textbf{DeepSeekV3} -- A frontier model from DeepSeek AI that pushes the boundaries of language understanding and generation. This model incorporates innovative architectural improvements and training methodologies, resulting in competitive performance on standard benchmarks. \textit{We note that according to official documentation and our preliminary testing, the current version of DeepSeekV3 exhibits inconsistent stability in tool-calling functionalities. This limitation was carefully accounted for in our experimental design and subsequent analysis to ensure fair comparisons across models.} The model api we use is: ``deepseek-chat''.
    
    \item \textbf{Qwen-2.5-Max} -- Alibaba's flagship model representing the pinnacle of their LLM research, featuring extensive pretraining on diverse multilingual corpora. The model demonstrates exceptional capabilities in both Chinese and English language processing, with impressive performance on complex reasoning, knowledge retrieval, and creative generation tasks. Its balanced capabilities across domains make it particularly valuable for evaluating the cross-lingual generalizability of our proposed methods. The model api we use is: ``qwen-max-0125''.
    
    \item \textbf{Gemini-2.0-Flash} -- Google's optimized model designed to balance computational efficiency with state-of-the-art performance. \textit{Our experimental design initially incorporated Gemini-2.0-Pro; however, due to its experimental status at the time of our research and consequent stability issues encountered during preliminary testing, we strategically pivoted to the more stable Flash variant. This decision ensured consistent and reliable results throughout our extensive evaluation process while still benefiting from Google's advanced LLM architecture.} The Flash variant provides excellent performance-to-efficiency ratio for our complex evaluation scenarios. The model api we use is: ``gemini-2.0-flash''
    
    \item \textbf{Llama-3.3-70B} -- Meta's open-source large language model with 70 billion parameters, representing one of the most powerful publicly available models. This model incorporates advanced training techniques and architectural innovations, resulting in exceptional performance across reasoning, coding, and general language understanding benchmarks. As an open-source model, Llama-3.3-70B offers unique transparency advantages and provides an important reference point for comparing proprietary and open-source approaches in our evaluation framework. The model we use is from a proxy where the api is ``meta-llama/Llama-3.3-70B-Instruct''
\end{itemize}

\subsection{Details about LLM Agents' System Prompts}
We utilize a set of system prompts to define the role and internal working logic of all agents. System prompts contains necessary information in free-text format, including role definition, task specification, available tools and corresponding descriptions, an example workflow and other important requirements. The workflow example acts as a few-shot hint to guide the agent's workflow. We also insert self-reflection requirements at the end of each system prompt, guiding the agents check their work again before returning.

\begin{mdframed}[backgroundcolor=lightgray!20]
\label{stage2_prompts}
\vspace{3pt}
\textbf{System prompts of Task Analyzer}
\vspace{5pt}
\hrule
\begin{verbatim}
You are acting as an agent for selecting a dataset that best 
matches a human user's requirements. You are provided with a 
list of dataset 
descriptions: 
{description_path}, which is a json containing a list of 
dictionaries. Every dictionary contains following entries: 
["dataset name", "dataset description", "dataset_path"].

You have access to the tools:
read_files: This function reads a script file (such as a Python 
file) so you can understand its content. Use it to read 
dataloader template file to grasp the expected format for the 
dataloader class.

Here is the typical workflow you should follow:
1. Use read_files to read {description_path}, understand its 
content.
2. choose exactly one dataset that best matches the user's 
requirements. Remember, your choice should mainly base on 
"dataset description".
3. Return the chosen dataset's name, description, and 
dataset_path,so a downstream peer agent can know these 
information accurately.
4. include <end> to end the conversation.

IMPORTANT NOTE: If you think there realy is no dataset that 
meets the user's requirements, then return no dataset. You
must always reflect on your choice and return reasons for 
your choice before ending.
\end{verbatim}
\end{mdframed}

\begin{mdframed}[backgroundcolor=lightgray!20]
\label{stage2_prompts}
\vspace{3pt}
\textbf{System prompts of Data Engineer}
\vspace{5pt}
\hrule
\begin{verbatim}
You are acting as an agent for preparing training and testing 
data in a clinical radiology context. I provide you with a raw, 
unprocessed dataset and its corresponding description, which 
can be found in {selector_content}. Your mission is to generate 
three files—train.json, test.json, and label_dict.json(if 
needed)—and save them to the working directory: {save_path}.
And you must make sure that the format of the json files matches 
some example files which will be mentioned below. Do not modify 
the original data files directly.
IMPORTANT NOTE: In selector content, you should be able to 
identify the dataset's name, the dataset description, and the 
dataset's root path.

You have access to the following tools:

list_files_in_second_level:
Use this tool to inspect the dataset's structure and determine 
where the image data files are located (typically large files 
such as png, jpg, nii, pck, pt, npy, etc.) along with any 
metadata or label files (usually csv, json, txt, etc.). This 
help you decide which files to read further.
preview_file_content:
Some metadata or label files might be very large and exceed 
the context length. Use this tool to preview a portion of the 
file so that you understand its structure well enough to plan 
how to parse it using code.
write_file:
Your final goal is to generate train.json, test.json, and 
label_dict.json(if needed). Therefore, you should write a Python 
script that creates these files. Once you fully understand the 
folder structure and the content of the metadata/label files, 
write code (possibly using libraries such as os, re, pandas, 
etc.) to traverse or read the metadata files and extract the 
needed entries for each data item.
read_files: 
This function reads a script file (such as a Python file) so you 
can understand its content. Use it to read the dataloader 
template file to grasp the expected format for the dataloader 
class. Pay attention, you can not apply this tool to read 
train/test.json.
edit_file:
Since errors might occur during your first code attempt, you can 
use this tool to modify your files based on error messages or 
feedback.
run_script:
Use this tool to execute your Python file from the command line. 
Remember that this is the only tool available to run code.
Here is the typical workflow you should follow:

Based on the dataset's description, use the 
list_files_in_second_level tool to understand the organization 
and structure of the dataset. Identify files that are likely 
to contain metadata or labels.
Use the preview_file_content tool to read a portion of these 
files so that you can understand their structure and determine 
how to parse them with your code.
Based on the dataset's description, You Must Use the 
traverse_dirs tool and read_files tool to read the directory 
structure of {examples_path}, and find the example output jsons 
based on the medical task, for the next step to refer to.
Once you feel that you have a sufficient understanding, write a 
Python script under director {save_path} (using the write_file 
tool) that generates the following:
[train.json ,test.json and label_dict.json(if needed)]
Remember, If label_dict.json is not provided by chosen example 
files, then you must not generate it!!!

IMPORTANT: You Must Make sure that the json files you output 
matches the format of your chosen example files! Especially the 
dictionary structure!
If you wan to read a file named 'labels.json', use read_files 
instead of preview_file_content!

train/test split: Ensure that the data is split into training 
and testing sets in a reasonable ratio (e.g., 80/20) and that 
the split is random. If train/test split is already presented, 
you don't need to split, but you still need to generate the 
json files. Besides, ensures that for each training and testing 
sample the key-value pairs in the dictionary are internally 
shuffled.
Use the run_script tool to execute your script. If errors occur 
during execution, you can use the edit_file tool to modify your 
code until the script runs successfully and produces the three 
JSON files.
Remember, your objective is to automate the creation of shuffled 
train.json, test.json, and label_dict.json(if needed) without 
altering the raw data files directly.
Remember, the formats of train.json, test.json and 
label_dict.json(if exists) must follow the example files.
Before ending, you should reflect on your work. If you think 
there is no error anymore and all the json files are generated, 
please conclude your work and include <end> to end the 
conversations.
\end{verbatim}
\end{mdframed}

\begin{mdframed}[backgroundcolor=lightgray!20]
\label{stage2_prompts}
\vspace{3pt}
\textbf{System prompts of Module Architect}
\vspace{5pt}
\hrule
\begin{verbatim}
You are acting as an agent responsible for writing a dataloader 
for a dataset. Your ultimate goal is to create a 'dataloader.py' 
script that will be used to feed data into the training process 
under the {dataindex_path}. The dataset index files are located 
at dataindex_path: {dataindex_path} and contain train.json, 
test.json, and label_dict.json(may not exist). You must also 
choose a template file located at {template_path}, refer to it.

A peer dataset processor has already generated these index 
files, (informations can be found in {processor_msg}) so your 
task is to write a dataloader class that can read these files 
and load the data into the training process. The dataloader 
should be able to handle the training and testing data, as well 
as the label dictionary.

Datast description is also provided in: {description}

You have access to a series of utility functions, which are as 
follows:

traverse_dirs: 
This tool traverses a given directory and returns all the file 
paths under that directory. It helps you understand the folder 
structure—in this case, to inspect the various files in the 
index folder.
preview_file_content: 
Often, JSON files have extensive content that might exceed the 
context 
length. Use this function to quickly understand the structure 
of the JSON files (train.json, test.json, label_dict.json) so 
you know how to structure your dataloader class.
read_file: 
This function reads a script file (such as a Python file) so 
you can understand its content. Use it to read the dataloader 
template file to grasp the expected format for the dataloader 
class. Pay attention, you 
MUST NOT apply this tool to read train/test.json.
write_file: 
This utility writes a file, typically a Python script. After 
reviewing the template and understanding the structures of 
train, test, and label_dict(if exists), use this tool to 
create your dataloader class and save it as dataloader.py for 
training data ingestion.
edit_file: 
If the initial write_file output encounters issues, you can 
use edit_file to modify the script based on error messages or 
additional prompts.
run_script: 
Once the dataloader is written, use this tool to execute the 
script and verify that it correctly loads the data without 
errors.
A sample workflow might be:

Directory Inspection: Use traverse_dirs to read the directory 
structure of the given path {dataindex_path}, identifying the 
presence of the train, test, and label_dict(may not exist) 
JSON index files.Preview JSON Content: Employ 
preview_file_content to inspect these JSON files and understand 
their structures.
Choose Template: Based on the medical task, which can be found 
in dataset description, choose a dataloader code template from 
{template_path} for reference.
Review Template: Utilize read_file to examine your chosen 
dataloader template file and understand the proper format for 
writing the dataloader class. Remember that this is not the 
end! you must go on to write dataloader.py
Code Development: Based on the insights from the JSON structures 
and the 
template, use write_file to write your dataloader class to 
'{dataindex_path}/dataloader.py' (Include a main function if 
necessary).
Save and Test: After writing dataloader.py, you must use 
run_script to test and verify that the script runs correctly!!! 
You must put your dataloader.py under {dataindex_path}!!!
Debug and Validate: If errors occur, use edit_file and 
run_script as needed to debug the script until it fully 
processes the entire dataset.

Remember: Your task is to write & validate a dataloader that 
successfully iterates over the dataset and verifies that it 
runs correctly during training. Always refer to the template 
for guidance on the expected format.

You MUST use write_file to create a dataloader.py under 
{dataindex_path} and verify that it runs correctly!!!
You MUST tell where you place dataloader.py!!!
You should write dataloader according only to the json files 
and the template. And try not to modify too much of the 
template. If you see comments in the template like "you must 
not modify this line", then do not modify it.

If you think your dataloader.py is ready, and the dataloader 
is already validated, please conclude your work and include 
<end> to end the conversation.

Important: When you use write_file tool, print the parameters 
you pass to the tool function!!!
Before ending, you should reflect on your work. If you think 
there is no error anymore and all the json files are generated, 
please conclude your work and include <end> to end the 
conversations.
\end{verbatim}
\end{mdframed}

\begin{mdframed}[backgroundcolor=lightgray!20]
\label{stage2_prompts}
\vspace{3pt}
\textbf{System prompts of Model Trainer}
\vspace{5pt}
\hrule
\begin{verbatim}
You are an AI assistant specialized in radiology tasks, 
capable of writing training code, executing training processes, 
and debugging. 
Your primary focus areas include disease diagnosis, organ 
segmentation, anomaly detection, and report generation tasks. 
You handle end-to-end code writing, debugging, and training.

peer processor and dataloader agents have completed preliminary 
tasks of dataset preparation and dataloader class writing, 
messages documented in {processor_msg} and {dataloader_msg}. 
You will build upon this groundwork.

Your working directory is {work_path}, all operations must be 
strictly confined to this directory. To accomplish training 
tasks, you have access to the following tools:

1. traverse_dirs: Used for recursively traversing file paths in 
the workspace to understand directory structure and infer file 
purposes from their names.
2. read_files: Used to read content from one or multiple files 
to understand implementation details and determine if changes or 
operations are needed. Please avoid read datapath files such as 
the train.json, test.json and label_dict.json.
3. write_file: Used for implementing new functional code.
4. edit_file: Used for modifying files, including adding data 
and model information to template files, adding new features, 
or fixing errors. 
Note that when using this tool to edit a file, please always 
firstly read the content before.
5. run_script: Used for executing training through sh scripts.
6. copy_files: Used to copy a file to a new path, typically 
used when copying train.py and train.sh from ReferenceFiles to 
workspace

You can also access {train_script_path} to choose and copy the 
best matching train.py and train.sh to workspace. But you cannot 
edit files under {train_script_path}.

Important notes:
- The Datapath, Loss, and Utils directories respectively contain 
JSON/csv/JSONL data indices for training/validation and dataset 
class you need, loss functions, and utility packages. While 
these shouldn't be modified, you must understand their 
relationships and functions.
- The Logout directory stores training results and should not be 
manually written to.
- The Model directory contains training code modules for 
different tasks. Generally, these shouldn't be modified, but you 
should read them to understand their functionality and usage. 
Remember that if the medical task is Organ Segmnentation, you do 
not have to read Modeldirectory, because model is provided in 
train.py already.
- The directory {train_script_path} contains different medical 
tasks' respective train.sh and train.py files, you should choose 
the best matching train.sh and train.py based on medical task, 
and copy them to workspace.
- train.py contains the main training code template using 
transformers trainer framework. You need to carefully read and 
modify its contents as needed.
- train.sh is the script for running the main code, containing 
parameter settings that you need to understand and configure.
- train.py has some code lines commented by sth like 'you should 
not modify this line', if you see this, don't modify that line.

The workflow consists of three phases:
1. traverse {train_script_path} to choose the best matching 
train.sh and train.py based on medical task, and use copy_files 
to copy to workspace.
2. Understanding structure and reading files/code templates
3. Initial code adjustment and refinement. Modify train.py and 
train.sh to make them ready. A Hint: You always have to import 
the dataset class from {work_path}/Datapath/dataloader.py
4. Script execution (use run_script tool to execute train.sh) 
and debug loop until successful training completion

Phase 1 requires traversing train_script_path, choosing and 
copying the best train.sh and train.py to workspace.
Phase 2 requires traversing the working directory and reading 
all crucial code to understand their connections. 
Phase 3 involves careful review of train.py and train.sh, 
making necessary modifications to achieve an executable version. 
Phase 4 involves executing train.sh and iteratively fixing 
errors based on error messages until successful execution.

IMPORTANT: You must execute train.sh and make sure it's running 
normally before you exit

Before each operation, you should consider its purpose and 
verify its appropriateness, especially when uncertain or 
experiencing potential hallucinations. Use traverse or read 
tools to check and understand corresponding parts. Always 
remember your final goal is to successfully run the training 
script.
Before ending, you should reflect on your work. If you think 
there is no error anymore and all the json files are generated, 
please conclude your work and include <end> to end the 
conversations.
\end{verbatim}
\end{mdframed}

\subsection{Details about Prompt for Comparison Experiments}
For single anget systems such as ML-AgentBench, Aider, Cursor Composer, Windsurf Cascade and Github Copilot Edits. We use a prompt combining four role-specific agents' prompt as below:

\begin{mdframed}[backgroundcolor=lightgray!20]
\label{stage2_prompts}
\vspace{3pt}
\textbf{System Prompt for Single Agent}
\vspace{5pt}
\hrule
\begin{verbatim}

End-to-End Machine Learning Pipeline Agent Prompt
Objective
Build an end-to-end machine learning pipeline that includes:
Dataset selection and processing: Choose the dataset that best 
fits the user's requirements.
JSON index generation: Create train.json, test.json, and (when 
applicable) label_dict.json files without modifying any raw 
data.
Dataloader development: Write a dataloader.py script to feed 
data into the training process.
Training script preparation and execution: Select and prepare 
training scripts (train.sh and train.py), execute them, and 
ensure training runs successfully.
All operations must remain strictly within working directory.
Provided Paths
Dataset Description File:
description_path = "/path/to/DataCard/descriptions.json"
Save Path / Data Index Path:
save_path = "/path/to/TrainPipeline/Datapath"
 (dataindex_path = save_path)
Example JSON Files Directory:
examples_path = "/path/to/ReferenceFiles/DataJsonExamples"
Dataloader Template Directory:
template_path = "/path/to/ReferenceFiles/DataLoaderExamples"
Working Directory:
work_path = "/path/to/TrainPipeline"
Training Scripts Directory:
train_script_path = "/path/to/ReferenceFiles/TrainingScripts"
Phase 1: Dataset Selection
Understanding the Dataset Descriptions:
Read the JSON file at {description_path} to view all dataset 
entries. Each dataset entry is a dictionary with keys: 
["dataset name", "dataset description", "dataset_path"].
Dataset Choice:
Action: Choose exactly one dataset that best fits the user's 
requirements, basing the decision primarily on the "dataset 
description" entry.
Outcome: Return the chosen dataset’s name, dataset description, 
and dataset_path so that a downstream peer agent receives this 
information accurately.
Note: If no dataset fulfills the user's requirements, provide 
reasons and end the conversation by outputting <end>.
Phase 2: JSON Index Generation
Examination of the Dataset Structure:
Inspect the dataset directory structure using directory 
traversal methods.
Identify files that likely contain metadata or labels by 
previewing a portion of their contents.
Additionally, review the example outputs in {examples_path} to 
understand the expected JSON format for the medical task.
Creating the Splitting Script:
Objective: Write a Python script (to be saved under {save_path}) 
that generates the following files: 
train.json, test.json, label_dict.json (only if such a file is 
provided in the examples—the file should not be generated
otherwise)
Data Splitting:
If the raw dataset already provides a train/test split, simply 
reformat and output the JSONs. Otherwise, perform an 80/20 
random split.
Additional Requirement: For every sample in the JSON files, the 
key-value pairs within each dictionary should be randomly 
shuffled.
Hint: When you need to inspect file content or directory 
structures during development, invoke the appropriate file 
inspection functions.
Action: Write the full Python script accordingly, then test it 
by executing the script.
Debugging: Modify the code as needed until it runs without 
errors.
Completion: Once the JSON files are successfully generated, 
append <end> to indicate the phase is complete.
Phase 3: Dataloader Creation
Inspecting the Data Indices:
Traverse the directory at {dataindex_path} to confirm the 
presence of train.json, test.json, and (if it exists) 
label_dict.json.
Preview their content to understand the JSON structure.
Selecting a Dataloader Template:
From the directory {template_path}, select the dataloader code 
template that best aligns with the medical task (as described 
in the dataset description).
Read the chosen template thoroughly to grasp its format and any 
constraints (for example, lines with comments such as “you must 
not modify this line” must remain unchanged).
Writing the Dataloader:
Task: Develop dataloader.py to load and iterate over the 
training and testing data, handling the label dictionary if 
available.
Implementation: Write the code based on the insights from the 
JSON structure and the dataloader template.
Save Location: Place dataloader.py under {dataindex_path}.
Testing: Execute the dataloader to ensure it runs correctly. 
Make any necessary adjustments by editing the file.
Reporting: Clearly indicate where dataloader.py has been placed.
Completion: When the dataloader is functioning properly, 
include <end>.
Phase 4: Training Script Preparation and Execution
Phase Overview:
Goal: Set up and run a training process using the chosen 
training scripts for the specific medical task.
Script Selection:
Traverse {train_script_path} to find the best matching train.sh 
and train.py based on the medical task requirements.
Action: Copy the selected files into the workspace.
Code Review and Integration:
Review the directory structure: 
Datapath: Contains the JSON indices.
Loss/Utils: Contains needed loss functions and utility packages.
Model: Contains model modules (for Organ Segmentation tasks, 
this may already be provided).
Update train.py and train.sh as needed. They must import the 
dataset class from {work_path}/Datapath/dataloader.py.
Caution: Do not modify any lines explicitly marked “you must 
not modify this line.”
Execution and Debugging:
Action: Run the training script by executing train.sh.
If errors occur, modify the scripts iteratively until training 
executes successfully.
Final Check: Ensure that train.sh is running normally.
Completion:
Once the training script is validated and functions as intended,
mark the phase completion by including <end>.
Key Focus Areas
Disease Diagnosis
Organ Segmentation
Anomaly Detection
Report Generation Tasks
Critical Reminders
Operation Boundaries: All operations must remain confined to 
the working directory ({work_path}).
When to Call Tools vs. Write Code:
Inspecting files or traversing directories? Use file inspection 
functions.
When generating or modifying code? Write or edit the code 
directly.
Do Not Alter Raw Data: Always generate derived files (such as 
JSON indices or scripts) in the appropriate directories.
Validation is Crucial: Continuously test your development steps 
and ensure scripts run correctly before moving on.

\end{verbatim}
\end{mdframed}

\subsection{Details about the Agent system comparison Experiments}
Here we run each task experiments twice, resulting in a 28 execution in total. Here we reorganize them categorized by radiology task-level: Organ Segmentation, Anomaly Detection, Disease Diagnosis and Report Generation. We detail our each execution under each agentic framework, using metrics including Average Actions and Iterarions where one action means a step of tool using and a iteration means a step of debug for script execution. The result are shown as below:

\begin{table}[]
\caption{Performance metrics of ML-AgentBench across various medical imaging tasks.}
\label{tab:ml-agentbench}
\centering
\footnotesize
\begin{tabular}{l|l|ccc|ccc}
\toprule
\multicolumn{2}{l|}{\textbf{ML-AgentBench}} & \multicolumn{3}{c|}{\textbf{Execution 1}} & \multicolumn{3}{c}{\textbf{Execution 2}} \\ 
\cmidrule(r){3-5} \cmidrule(r){6-8}
\multicolumn{2}{l|}{} & \textbf{succ} & \textbf{actions} & \textbf{debug loops} & \textbf{succ} & \textbf{actions} & \textbf{debug loops} \\ 
\midrule
\multirow{4}{*}{DisDiag} & ADNI & 0 & - & - & 0 & - & - \\ 
 & KneeMRI & 0 & - & - & 0 & - & - \\ 
 & CC-CCII & 0 & - & - & 0 & - & - \\ 
 & CT-Kidney & 0 & - & - & 0 & - & - \\ 
\midrule
\multirow{4}{*}{AnoDet} & COVID19 & 0 & - & - & 0 & - & - \\ 
 & INSTANCE2022 & 0 & - & - & 0 & - & - \\ 
 & MSD Pancreas & 0 & - & - & 0 & - & - \\ 
 & ChestX-Det10 & 0 & - & - & 0 & - & - \\ 
\midrule
\multirow{3}{*}{RepGene} & CT-RATE & 0 & - & - & 0 & - & - \\ 
 & BrainGnome & 0 & - & - & 0 & - & - \\ 
 & IU\_Xray & 0 & - & - & 0 & - & - \\ 
\midrule
\multirow{3}{*}{OrgSeg} & BTCV & 0 & - & - & 0 & - & - \\ 
 & VerSe & 0 & - & - & 0 & - & - \\ 
 & L2R-OASIS & 0 & - & - & 0 & - & - \\ 
\bottomrule
\end{tabular}
\end{table}

\begin{table}[]
\caption{Performance metrics of Aider across various medical imaging tasks.}
\label{tab:aider}
\centering
\footnotesize
\begin{tabular}{l|l|ccc|ccc}
\toprule
\multicolumn{2}{l|}{\textbf{Aider}} & \multicolumn{3}{c|}{\textbf{Execution 1}} & \multicolumn{3}{c}{\textbf{Execution 2}} \\ 
\cmidrule(r){3-5} \cmidrule(r){6-8}
\multicolumn{2}{l|}{} & \textbf{succ} & \textbf{actions} & \textbf{debug loops} & \textbf{succ} & \textbf{actions} & \textbf{debug loops} \\ 
\midrule
\multirow{4}{*}{DisDiag} & ADNI & 0 & - & - & 0 & - & - \\ 
 & KneeMRI & 1 & 48 & 4 & 0 & - & - \\ 
 & CC-CCII & 0 & - & - & 0 & - & - \\ 
 & CT-Kidney & 0 & - & - & 0 & 41 & 5 \\ 
\midrule
\multirow{4}{*}{AnoDet} & COVID19 & 0 & 58 & 8 & 0 & 42 & 5 \\ 
 & INSTANCE2022 & 0 & - & - & 0 & - & - \\ 
 & MSD Pancreas & 0 & - & - & 0 & 42 & 5 \\ 
 & ChestX-Det10 & 0 & - & - & 0 & - & - \\ 
\midrule
\multirow{3}{*}{RepGene} & CT-RATE & 0 & 61 & 6 & 0 & - & - \\ 
 & BrainGnome & 0 & - & - & 0 & 58 & 6 \\ 
 & IU\_Xray & 0 & - & - & 0 & 51 & 5 \\ 
\midrule
\multirow{3}{*}{OrgSeg} & BTCV & 0 & - & - & 0 & - & - \\ 
 & VerSe & 0 & 67 & 7 & 0 & 49 & 6 \\ 
 & L2R-OASIS & 0 & - & - & 0 & - & - \\ 
\bottomrule
\end{tabular}
\end{table}

\begin{table}[]
\caption{Performance metrics of Cursor Composer across various medical imaging tasks.}
\label{tab:cursor-composer}
\centering
\footnotesize
\begin{tabular}{l|l|ccc|ccc}
\toprule
\multicolumn{2}{l|}{\textbf{Cursor Composer}} & \multicolumn{3}{c|}{\textbf{Execution 1}} & \multicolumn{3}{c}{\textbf{Execution 2}} \\ 
\cmidrule(r){3-5} \cmidrule(r){6-8}
\multicolumn{2}{l|}{} & \textbf{succ} & \textbf{actions} & \textbf{debug loops} & \textbf{succ} & \textbf{actions} & \textbf{debug loops} \\ 
\midrule
\multirow{4}{*}{DisDiag} & ADNI & 0 & 27 & 3 & 0 & - & - \\ 
 & KneeMRI & 0 & - & - & 1 & 44 & 5 \\ 
 & CC-CCII & 0 & - & - & 0 & - & - \\ 
 & CT-Kidney & 0 & 38 & 4 & 0 & - & - \\ 
\midrule
\multirow{4}{*}{AnoDet} & COVID19 & 0 & - & - & 0 & 29 & 4 \\ 
 & INSTANCE2022 & 1 & - & - & 0 & - & - \\ 
 & MSD Pancreas & 0 & 41 & 6 & 1 & 37 & 3 \\ 
 & ChestX-Det10 & 1 & - & - & 0 & - & - \\ 
\midrule
\multirow{3}{*}{RepGene} & CT-RATE & 0 & - & - & 0 & - & - \\ 
 & BrainGnome & 0 & 54 & 6 & 0 & - & - \\ 
 & IU\_Xray & 0 & 49 & 5 & 0 & - & - \\ 
\midrule
\multirow{3}{*}{OrgSeg} & BTCV & 0 & - & - & 1 & 42 & 7 \\ 
 & VerSe & 1 & - & - & 0 & - & - \\ 
 & L2R-OASIS & 0 & - & - & 0 & - & - \\ 
\bottomrule
\end{tabular}
\end{table}

\begin{table}[]
\caption{Performance metrics of Windsurf Cascade across various medical imaging tasks.}
\label{tab:windsurf-cascade}
\centering
\footnotesize
\begin{tabular}{l|l|ccc|ccc}
\toprule
\multicolumn{2}{l|}{\textbf{Windsurf Cascade}} & \multicolumn{3}{c|}{\textbf{Execution 1}} & \multicolumn{3}{c}{\textbf{Execution 2}} \\ 
\cmidrule(r){3-5} \cmidrule(r){6-8}
\multicolumn{2}{l|}{} & \textbf{succ} & \textbf{actions} & \textbf{debug loops} & \textbf{succ} & \textbf{actions} & \textbf{debug loops} \\ 
\midrule
\multirow{4}{*}{DisDiag} & ADNI & 0 & 29 & 5 & 0 & 32 & 4 \\ 
 & KneeMRI & 0 & 41 & 5 & 0 & - & - \\ 
 & CC-CCII & 0 & - & - & 0 & - & - \\ 
 & CT-Kidney & 0 & - & - & 0 & 38 & 5 \\ 
\midrule
\multirow{4}{*}{AnoDet} & COVID19 & 0 & - & - & 0 & - & - \\ 
 & INSTANCE2022 & 0 & - & - & 0 & - & - \\ 
 & MSD Pancreas & 0 & 32 & 4 & 0 & 36 & 4 \\ 
 & ChestX-Det10 & 0 & - & - & 0 & 40 & 5 \\ 
\midrule
\multirow{3}{*}{RepGene} & CT-RATE & 0 & - & - & 0 & - & - \\ 
 & BrainGnome & 0 & 48 & 5 & 0 & - & - \\ 
 & IU\_Xray & 0 & - & - & 0 & 49 & 6 \\ 
\midrule
\multirow{3}{*}{OrgSeg} & BTCV & 0 & - & - & 0 & 39 & 5 \\ 
 & VerSe & 0 & - & - & 0 & - & - \\ 
 & L2R-OASIS & 0 & - & - & 0 & - & - \\ 
\bottomrule
\end{tabular}
\end{table}

\begin{table}[]
\caption{Performance metrics of Copilot Edits across various medical imaging tasks.}
\label{tab:copilot-edits}
\centering
\footnotesize
\begin{tabular}{l|l|ccc|ccc}
\toprule
\multicolumn{2}{l|}{\textbf{Copilot Edits}} & \multicolumn{3}{c|}{\textbf{Execution 1}} & \multicolumn{3}{c}{\textbf{Execution 2}} \\ 
\cmidrule(r){3-5} \cmidrule(r){6-8}
\multicolumn{2}{l|}{} & \textbf{succ} & \textbf{actions} & \textbf{debug loops} & \textbf{succ} & \textbf{actions} & \textbf{debug loops} \\ 
\midrule
\multirow{4}{*}{DisDiag} & ADNI & 0 & - & - & 0 & - & - \\ 
 & KneeMRI & 0 & 44 & 4 & 0 & - & - \\ 
 & CC-CCII & 0 & - & - & 0 & 39 & 3 \\ 
 & CT-Kidney & 0 & - & - & 0 & 51 & 5 \\ 
\midrule
\multirow{4}{*}{AnoDet} & COVID19 & 0 & 46 & 3 & 0 & 41 & 3 \\ 
 & INSTANCE2022 & 0 & - & - & 0 & - & - \\ 
 & MSD Pancreas & 0 & 55 & 5 & 0 & - & - \\ 
 & ChestX-Det10 & 0 & 39 & 3 & 0 & - & - \\ 
\midrule
\multirow{3}{*}{RepGene} & CT-RATE & 0 & - & - & 0 & 52 & 4 \\ 
 & BrainGnome & 0 & - & - & 0 & - & - \\ 
 & IU\_Xray & 0 & 41 & 4 & 0 & - & - \\ 
\midrule
\multirow{3}{*}{OrgSeg} & BTCV & 0 & - & - & 0 & - & - \\ 
 & VerSe & 0 & 51 & 4 & 0 & 45 & 3 \\ 
 & L2R-OASIS & 0 & - & - & 0 & - & - \\ 
\bottomrule
\end{tabular}
\end{table}

\begin{table}[]
\caption{Performance metrics of Ours across various medical imaging tasks.}
\label{tab:ours}
\centering
\footnotesize
\begin{tabular}{l|l|ccc|ccc}
\toprule
\multicolumn{2}{l|}{\textbf{Ours}} & \multicolumn{3}{c|}{\textbf{Execution 1}} & \multicolumn{3}{c}{\textbf{Execution 2}} \\ 
\cmidrule(r){3-5} \cmidrule(r){6-8}
\multicolumn{2}{l|}{} & \textbf{succ} & \textbf{actions} & \textbf{debug loops} & \textbf{succ} & \textbf{actions} & \textbf{debug loops} \\ 
\midrule
\multirow{4}{*}{DisDiag} & ADNI & 0 & 48 & 4 & 0 & 32 & 2 \\ 
 & KneeMRI & 0 & 43 & 3 & 0 & 31 & 1 \\ 
 & CC-CCII & 0 & 43 & 2 & 0 & 26 & 1 \\ 
 & CT-Kidney & 0 & 27 & 1 & 0 & 31 & 2 \\ 
\midrule
\multirow{4}{*}{AnoDet} & COVID19 & 0 & 29 & 2 & 0 & 28 & 2 \\ 
 & INSTANCE2022 & 0 & 34 & 3 & 0 & 38 & 3 \\ 
 & MSD Pancreas & 0 & - & - & 0 & 27 & 2 \\ 
 & ChestX-Det10 & 0 & 31 & 3 & 0 & 30 & 2 \\ 
\midrule
\multirow{3}{*}{RepGene} & CT-RATE & 0 & 29 & 2 & 0 & - & - \\ 
 & BrainGnome & 0 & 38 & 4 & 0 & - & - \\ 
 & IU\_Xray & 0 & 31 & 1 & 0 & 30 & 2 \\ 
\midrule
\multirow{3}{*}{OrgSeg} & BTCV & 0 & 29 & 1 & 0 & 30 & 3 \\ 
 & VerSe & 0 & 34 & 1 & 0 & - & - \\ 
 & L2R-OASIS & 0 & - & - & 0 & 35 & 2 \\ 
\bottomrule
\end{tabular}
\end{table}

\begin{table}[]
\caption{Performance metrics of the Execution without collaboration (single agent) across various medical imaging tasks.}
\label{tab:wo-colab}
\centering
\footnotesize
\begin{tabular}{l|l|ccc|ccc}
\toprule
\multicolumn{2}{l|}{\textbf{w/o Colab}} & \multicolumn{3}{c|}{\textbf{Execution 1}} & \multicolumn{3}{c}{\textbf{Execution 2}} \\ 
\cmidrule(r){3-5} \cmidrule(r){6-8}
\multicolumn{2}{l|}{} & \textbf{succ} & \textbf{actions} & \textbf{debug loops} & \textbf{succ} & \textbf{actions} & \textbf{debug loops} \\ 
\midrule
\multirow{4}{*}{DisDiag} & ADNI & 0 & 34 & 4 & 0 & - & - \\ 
 & KneeMRI & 0 & 37 & 5 & 0 & - & - \\ 
 & CC-CCII & 0 & - & - & 0 & - & - \\ 
 & CT-Kidney & 0 & - & - & 0 & 29 & 4 \\ 
\midrule
\multirow{4}{*}{AnoDet} & COVID19 & 0 & 40 & 5 & 0 & - & - \\ 
 & INSTANCE2022 & 0 & 36 & 4 & 0 & - & - \\ 
 & MSD Pancreas & 0 & - & - & 0 & 33 & 5 \\ 
 & ChestX-Det10 & 0 & 41 & 5 & 0 & - & - \\ 
\midrule
\multirow{3}{*}{RepGene} & CT-RATE & 0 & - & - & 0 & - & - \\ 
 & BrainGnome & 0 & 32 & 5 & 0 & - & - \\ 
 & IU\_Xray & 0 & - & - & 0 & 35 & 4 \\ 
\midrule
\multirow{3}{*}{OrgSeg} & BTCV & 0 & - & - & 0 & 33 & 5 \\ 
 & VerSe & 0 & 36 & 5 & 0 & - & - \\ 
 & L2R-OASIS & 0 & 32 & 4 & 0 & - & - \\ 
\bottomrule
\end{tabular}
\end{table}

\begin{table}[]
\caption{Performance metrics of the Execution without extensive debugging across various medical imaging tasks.}
\label{tab:wo-debug}
\centering
\footnotesize
\begin{tabular}{l|l|ccc|ccc}
\toprule
\multicolumn{2}{l|}{\textbf{w/o debug}} & \multicolumn{3}{c|}{\textbf{Execution 1}} & \multicolumn{3}{c}{\textbf{Execution 2}} \\ 
\cmidrule(r){3-5} \cmidrule(r){6-8}
\multicolumn{2}{l|}{} & \textbf{succ} & \textbf{actions} & \textbf{debug loops} & \textbf{succ} & \textbf{actions} & \textbf{debug loops} \\ 
\midrule
\multirow{4}{*}{DisDiag} & ADNI & 0 & - & - & 0 & - & - \\ 
 & KneeMRI & 0 & - & - & 1 & 33 & 1 \\ 
 & CC-CCII & 0 & - & - & 0 & - & - \\ 
 & CT-Kidney & 0 & - & - & 0 & - & - \\ 
\midrule
\multirow{4}{*}{AnoDet} & COVID19 & 0 & - & - & 0 & - & - \\ 
 & INSTANCE2022 & 1 & 22 & 1 & 0 & - & - \\ 
 & MSD Pancreas & 0 & - & - & 1 & 26 & 1 \\ 
 & ChestX-Det10 & 1 & 24 & 1 & 0 & - & - \\ 
\midrule
\multirow{3}{*}{RepGene} & CT-RATE & 0 & - & - & 0 & - & - \\ 
 & BrainGnome & 0 & - & - & 0 & - & - \\ 
 & IU\_Xray & 0 & - & - & 0 & - & - \\ 
\midrule
\multirow{3}{*}{OrgSeg} & BTCV & 0 & - & - & 1 & 31 & 1 \\ 
 & VerSe & 1 & 30 & 1 & 0 & - & - \\ 
 & L2R-OASIS & 0 & - & - & 0 & - & - \\ 
\bottomrule
\end{tabular}
\end{table}

\begin{table}[]
\caption{Performance metrics of the Execution without reflection mechanism across various medical imaging tasks.}
\label{tab:wo-reflect}
\centering
\footnotesize
\begin{tabular}{l|l|ccc|ccc}
\toprule
\multicolumn{2}{l|}{\textbf{w/o reflect}} & \multicolumn{3}{c|}{\textbf{Execution 1}} & \multicolumn{3}{c}{\textbf{Execution 2}} \\ 
\cmidrule(r){3-5} \cmidrule(r){6-8}
\multicolumn{2}{l|}{} & \textbf{succ} & \textbf{actions} & \textbf{debug loops} & \textbf{succ} & \textbf{actions} & \textbf{debug loops} \\ 
\midrule
\multirow{4}{*}{DisDiag} & ADNI & 1 & 37 & 4 & 0 & - & - \\ 
 & KneeMRI & 1 & 35 & 3 & 1 & 34 & 5 \\ 
 & CC-CCII & 1 & 31 & 4 & 1 & 36 & 4 \\ 
 & CT-Kidney & 1 & 36 & 3 & 1 & 39 & 4 \\ 
\midrule
\multirow{4}{*}{AnoDet} & COVID19 & 1 & 25 & 5 & 1 & 23 & 5 \\ 
 & INSTANCE2022 & 1 & 20 & 3 & 1 & 23 & 4 \\ 
 & MSD Pancreas & 1 & 27 & 3 & 1 & 25 & 4 \\ 
 & ChestX-Det10 & 0 & - & - & 1 & 31 & 5 \\ 
\midrule
\multirow{3}{*}{RepGene} & CT-RATE & 1 & 42 & 6 & 1 & 46 & 7 \\ 
 & BrainGnome & 1 & 38 & 5 & 1 & 39 & 5 \\ 
 & IU\_Xray & 0 & - & - & 0 & - & - \\ 
\midrule
\multirow{3}{*}{OrgSeg} & BTCV & 1 & 28 & 2 & 1 & 27 & 2 \\ 
 & VerSe & 1 & 31 & 3 & 0 & - & - \\ 
 & L2R-OASIS & 0 & - & - & 1 & 27 & 2 \\ 
\bottomrule
\end{tabular}
\end{table}

\begin{table}[]
\caption{Performance metrics of the Execution without few-shot examples across various medical imaging tasks.}
\label{tab:wo-fewshot}
\centering
\footnotesize
\begin{tabular}{l|l|ccc|ccc}
\toprule
\multicolumn{2}{l|}{\textbf{w/o fewshot}} & \multicolumn{3}{c|}{\textbf{Execution 1}} & \multicolumn{3}{c}{\textbf{Execution 2}} \\ 
\cmidrule(r){3-5} \cmidrule(r){6-8}
\multicolumn{2}{l|}{} & \textbf{succ} & \textbf{actions} & \textbf{debug loops} & \textbf{succ} & \textbf{actions} & \textbf{debug loops} \\ 
\midrule
\multirow{4}{*}{DisDiag} & ADNI & 0 & - & - & 1 & 37 & 6 \\ 
 & KneeMRI & 1 & 31 & 4 & 1 & 29 & 3 \\ 
 & CC-CCII & 1 & 30 & 3 & 1 & 32 & 4 \\ 
 & CT-Kidney & 0 & - & - & 1 & 35 & 4 \\ 
\midrule
\multirow{4}{*}{AnoDet} & COVID19 & 1 & 21 & 3 & 0 & - & - \\ 
 & INSTANCE2022 & 0 & - & - & 0 & - & - \\ 
 & MSD Pancreas & 1 & 23 & 3 & 1 & 26 & 3 \\ 
 & ChestX-Det10 & 0 & - & - & 1 & 24 & 4 \\ 
\midrule
\multirow{3}{*}{RepGene} & CT-RATE & 0 & - & - & 0 & - & - \\ 
 & BrainGnome & 1 & 33 & 2 & 1 & 37 & 4 \\ 
 & IU\_Xray & 0 & - & - & 1 & 32 & 3 \\ 
\midrule
\multirow{3}{*}{OrgSeg} & BTCV & 0 & - & - & 1 & 39 & 5 \\ 
 & VerSe & 1 & 36 & 4 & 0 & - & - \\ 
 & L2R-OASIS & 1 & 37 & 3 & 0 & - & - \\ 
\bottomrule
\end{tabular}
\end{table}

\subsection{Details about the Agent Rols-specification Experiments.}
In the Role-specification analysis, we have mentioned that we run each task twice, which leads to 28 execution rounds in total. Here we reorganize them categorized by radiology task-level: Organ Segmentation, Anomaly Detection, Disease Diagnosis and Report Generation and in the following are each execution details:

\begin{table}[]
\caption{Performance metrics for different roles across tasks in OrganSeg, including Task Manager, Data Engineer, Module Architect, and Model Trainer.}
\label{tab:performance}
\centering
\footnotesize
\begin{tabular}{l|cccc|cccc|cccc|cccc}
\toprule
\textbf{OrgSeg} & \multicolumn{4}{c|}{\textbf{Task Manager}} & \multicolumn{4}{c|}{\textbf{Data Engineer}} & \multicolumn{4}{c|}{\textbf{Module Architect}} & \multicolumn{4}{c}{\textbf{Model Trainer}} \\ 
\cmidrule(r){2-5} \cmidrule(r){6-9} \cmidrule(r){10-13} \cmidrule(r){14-17}
 & \textbf{Run} & \textbf{Act} & \textbf{Iter} & \textbf{Tkn} & \textbf{Run} & \textbf{Act} & \textbf{Iter} & \textbf{Tkn} & \textbf{Run} & \textbf{Act} & \textbf{Iter} & \textbf{Tkn} & \textbf{Run} & \textbf{Act} & \textbf{Iter} & \textbf{Tkn} \\ 
\midrule
1 & 1(1) & 3 & 1 & 4k  & 1(1) & 8  & 1 & 49k  & 1(1) & 8  & 1 & 37k  & 1(1) & 10 & 3 & 139k \\ 
2 & 1(1) & 2 & 1 & 4k  & 1(1) & 12 & 1 & 85k  & 1(1) & 7  & 1 & 35k  & 1(1) & 9  & 2 & 75k  \\ 
3 & 1(1) & 2 & 1 & 5k  & 0(1) & 9  & 1 & 55k  & 1(1) & 12 & 3 & 88k  & 0(1) & 11 & 5 & 113k \\ 
4 & 1(1) & 2 & 1 & 4k  & 1(1) & 11 & 2 & 65k  & 0(1) & 10 & 2 & 60k  & 0(1) & 12 & 5 & 154k \\ 
5 & 1(1) & 2 & 1 & 5k  & 1(1) & 11 & 1 & 96k  & 1(1) & 10 & 2 & 62k  & 1(1) & 13 & 3 & 132k \\ 
6 & 1(1) & 2 & 1 & 4k  & 1(1) & 10 & 2 & 82k  & 1(1) & 16 & 5 & 164k & 1(1) & 9  & 2 & 78k  \\ 
\bottomrule
\end{tabular}
\end{table}

\begin{table}[]
\caption{Performance metrics for different roles across tasks in AnoDet, including Task Manager, Data Engineer, Module Architect, and Model Trainer.}
\label{tab:performance_anodet}
\centering
\footnotesize
\begin{tabular}{l|cccc|cccc|cccc|cccc}
\toprule
\textbf{AnoDet} & \multicolumn{4}{c|}{\textbf{Task Manager}} & \multicolumn{4}{c|}{\textbf{Data Engineer}} & \multicolumn{4}{c|}{\textbf{Module Architect}} & \multicolumn{4}{c}{\textbf{Model Trainer}} \\ 
\cmidrule(r){2-5} \cmidrule(r){6-9} \cmidrule(r){10-13} \cmidrule(r){14-17}
 & \textbf{Run} & \textbf{Act} & \textbf{Iter} & \textbf{Tkn} & \textbf{Run} & \textbf{Act} & \textbf{Iter} & \textbf{Tkn} & \textbf{Run} & \textbf{Act} & \textbf{Iter} & \textbf{Tkn} & \textbf{Run} & \textbf{Act} & \textbf{Iter} & \textbf{Tkn} \\ 
\midrule
1 & 1(1) & 2 & 1 & 4k  & 1(1) & 11 & 1 & 71k  & 1(1) & 9  & 1 & 43k  & 1(1) & 7  & 1 & 51k  \\ 
2 & 1(1) & 2 & 1 & 4k  & 1(1) & 10 & 1 & 60k  & 1(1) & 8  & 1 & 39k  & 1(1) & 8  & 2 & 60k  \\ 
3 & 1(1) & 2 & 1 & 5k  & 1(1) & 11 & 2 & 131k & 1(1) & 11 & 2 & 64k  & 1(1) & 10 & 3 & 73k  \\ 
4 & 1(1) & 2 & 1 & 4k  & 1(1) & 10 & 1 & 128k & 0(1) & 17 & 5 & 103k & 1(1) & 9  & 2 & 68k  \\ 
5 & 1(1) & 2 & 1 & 5k  & 1(1) & 11 & 2 & 71k  & 1(1) & 10 & 2 & 69k  & 0(1) & 13 & 5 & 92k  \\ 
6 & 1(1) & 2 & 1 & 4k  & 1(1) & 10 & 2 & 70k  & 1(1) & 10 & 3 & 72k  & 1(1) & 9  & 1 & 72k  \\ 
7 & 1(1) & 2 & 1 & 4k  & 1(1) & 8  & 1 & 107k & 1(1) & 9  & 1 & 45k  & 1(1) & 8  & 1 & 59k  \\ 
8 & 1(1) & 2 & 1 & 5k  & 1(1) & 9  & 1 & 99k  & 1(1) & 11 & 2 & 50k  & 1(1) & 10 & 2 & 62k  \\ 
\bottomrule
\end{tabular}
\end{table}

\begin{table}[]
\caption{Performance metrics for different roles across tasks in DisDiag, including Task Manager, Data Engineer, Module Architect, and Model Trainer.}
\label{tab:performance_disdiag}
\centering
\footnotesize
\begin{tabular}{l|cccc|cccc|cccc|cccc}
\toprule
\textbf{DisDiag} & \multicolumn{4}{c|}{\textbf{Task Manager}} & \multicolumn{4}{c|}{\textbf{Data Engineer}} & \multicolumn{4}{c|}{\textbf{Module Architect}} & \multicolumn{4}{c}{\textbf{Model Trainer}} \\ 
\cmidrule(r){2-5} \cmidrule(r){6-9} \cmidrule(r){10-13} \cmidrule(r){14-17}
 & \textbf{Run} & \textbf{Act} & \textbf{Iter} & \textbf{Tkn} & \textbf{Run} & \textbf{Act} & \textbf{Iter} & \textbf{Tkn} & \textbf{Run} & \textbf{Act} & \textbf{Iter} & \textbf{Tkn} & \textbf{Run} & \textbf{Act} & \textbf{Iter} & \textbf{Tkn} \\ 
\midrule
1 & 1(1) & 2 & 1 & 5k  & 1(1) & 7  & 1 & 153k & 1(1) & 9  & 1 & 39k  & 1(1) & 30 & 4 & 754k \\ 
2 & 1(1) & 2 & 1 & 4k  & 1(1) & 8  & 1 & 103k & 1(1) & 10 & 2 & 44k  & 1(1) & 13 & 2 & 95k  \\ 
3 & 1(1) & 2 & 1 & 4k  & 1(1) & 11 & 2 & 140k & 1(1) & 11 & 2 & 74k  & 1(1) & 19 & 4 & 330k \\ 
4 & 1(1) & 2 & 1 & 4k  & 1(1) & 9  & 1 & 112k & 1(1) & 10 & 2 & 69k  & 1(1) & 9  & 1 & 100k \\ 
5 & 1(1) & 2 & 1 & 5k  & 1(1) & 10 & 2 & 161k & 0(1) & 23 & 5 & 255k & 1(1) & 8  & 2 & 79k  \\ 
6 & 1(1) & 2 & 1 & 4k  & 1(1) & 8  & 1 & 92k  & 1(1) & 8  & 1 & 66k  & 1(1) & 8  & 1 & 79k  \\ 
7 & 1(1) & 2 & 1 & 4k  & 1(1) & 8  & 2 & 89k  & 1(1) & 10 & 2 & 81k  & 1(1) & 7  & 1 & 66k  \\ 
8 & 1(1) & 2 & 1 & 5k  & 1(1) & 10 & 1 & 77k  & 1(1) & 9  & 1 & 70k  & 1(1) & 10 & 1 & 82k  \\ 
\bottomrule
\end{tabular}
\end{table}

\begin{table}[]
\caption{Performance metrics for different roles across tasks in RepGene, including Task Analyzer, Data Engineer, Code Writer, and Model Trainer.}
\label{tab:performance_repgene}
\centering
\footnotesize
\begin{tabular}{l|cccc|cccc|cccc|cccc}
\toprule
\textbf{RepGene} & \multicolumn{4}{c|}{\textbf{Task Analyzer}} & \multicolumn{4}{c|}{\textbf{Data Engineer}} & \multicolumn{4}{c|}{\textbf{Code Writer}} & \multicolumn{4}{c}{\textbf{Model Trainer}} \\ 
\cmidrule(r){2-5} \cmidrule(r){6-9} \cmidrule(r){10-13} \cmidrule(r){14-17}
 & \textbf{Run} & \textbf{Act} & \textbf{Iter} & \textbf{Tkn} & \textbf{Run} & \textbf{Act} & \textbf{Iter} & \textbf{Tkn} & \textbf{Run} & \textbf{Act} & \textbf{Iter} & \textbf{Tkn} & \textbf{Run} & \textbf{Act} & \textbf{Iter} & \textbf{Tkn} \\ 
\midrule
1 & 1(1) & 2 & 1 & 4k  & 1(1) & 4  & 1 & 14k  & 1(1) & 5  & 1 & 18k  & 0(1) & 14 & 5 & 103k \\ 
2 & 1(1) & 2 & 1 & 5k  & 1(1) & 7  & 1 & 22k  & 1(1) & 10 & 2 & 33k  & 1(1) & 10 & 2 & 41k  \\ 
3 & 1(1) & 2 & 1 & 4k  & 1(1) & 11 & 1 & 94k  & 1(1) & 16 & 5 & 171k & 1(1) & 9  & 2 & 80k  \\ 
4 & 1(1) & 2 & 1 & 4k  & 1(1) & 10 & 1 & 80k  & 1(1) & 8  & 1 & 47k  & 1(1) & 11 & 1 & 62k  \\ 
5 & 1(1) & 2 & 1 & 4k  & 1(1) & 11 & 3 & 110k & 0(1) & 17 & 5 & 219k & 1(1) & 13 & 2 & 79k  \\ 
6 & 1(1) & 2 & 1 & 4k  & 1(1) & 9  & 1 & 77k  & 1(1) & 9  & 1 & 58k  & 1(1) & 10 & 1 & 55k  \\ 
\bottomrule
\end{tabular}
\end{table}

\end{document}